\documentclass[pdflatex,sn-mathphys-num]{sn-jnl}

\usepackage{graphicx}%
\usepackage{multirow}%
\usepackage{amsmath,amssymb,amsfonts}%
\usepackage{amsthm}%
\usepackage{mathrsfs}%
\usepackage[title]{appendix}%
\usepackage{xcolor}%
\usepackage{textcomp}%
\usepackage{manyfoot}%
\usepackage{booktabs}%
\usepackage{algorithm}%
\usepackage{algorithmicx}%
\usepackage{algpseudocode}%
\usepackage{listings}%
\usepackage{amssymb}
\usepackage{pifont}
\newcommand{\cmark}{\ding{51}}%
\newcommand{\xmark}{\ding{55}}%
\usepackage{tikz}
\usepackage{tablefootnote}

\usepackage[most]{tcolorbox}
\usepackage{enumitem}
\usepackage{lipsum}
\usepackage{listings}
\usepackage{xcolor}
\usepackage{parskip} 
\usepackage{subcaption}
\usepackage{caption}
\usepackage{siunitx}

\usepackage{tablefootnote}
\usepackage{hyperref}

\tcbuselibrary{listings, breakable, skins}

\definecolor{lightgray}{RGB}{245,245,245}
\definecolor{lightblue}{RGB}{230,245,255}
\definecolor{lightyellow}{RGB}{255,250,230}
\definecolor{lightgreen}{RGB}{240,255,240}
\definecolor{lightred}{RGB}{255, 204, 204}
\definecolor{lightpurple}{RGB}{245, 240, 255}
\definecolor{lightpink}{RGB}{255, 230, 236}
\definecolor{lightcyan}{RGB}{224, 245, 255}
\definecolor{lightorange}{RGB}{255, 243, 224}

\usetikzlibrary{arrows.meta, positioning}

\newcommand{\revisionpanos}[1]{\textcolor{black}{#1}}

\theoremstyle{thmstyleone}%
\theoremstyle{thmstyletwo}%

\theoremstyle{thmstylethree}%

\begin{document}

\title{A Systematic Evaluation of Large Language Models for PTSD Severity Estimation: The Role of Contextual Knowledge and Modeling Strategies}




\author*[1]{\fnm{Panagiotis} \sur{Kaliosis}}\email{pkaliosis@cs.stonybrook.edu}\equalcont{These authors contributed equally to this work.}

\author[1,2]{\fnm{Adithya} \sur{V Ganesan}}\equalcont{These authors contributed equally to this work.}

\author[2,3]{\fnm{Oscar N.E.} \sur{Kjell}}

\author[4]{\fnm{Whitney} \sur{Ringwald}}

\author[5]{\fnm{Scott} \sur{Feltman}}

\author[6]{\fnm{Melissa A.} \sur{Carr}}

\author[1]{\fnm{Dimitris} \sur{Samaras}}

\author[7]{\fnm{Camilo} \sur{Ruggero}}

\author[6,8]{\fnm{Benjamin J.} \sur{Luft}}

\author[9]{\fnm{Roman} \sur{Kotov}}

\author*[1,2]{\fnm{H. Andrew} \sur{Schwartz}}\email{hansen.schwartz@vanderbilt.edu}

\affil[1]{\orgdiv{Department of Computer Science}, \orgname{Stony Brook University}, \country{USA}}

\affil[2]{\orgdiv{College of Connected Computing}, \orgname{Vanderbilt University}, \country{USA}}

\affil[3]{\orgdiv{Department of Psychology}, \orgname{Lund University}, \country{Sweden}}

\affil[4]{\orgdiv{Department of Psychology}, \orgname{University of Minnesota}, \country{USA}}

\affil[5]{\orgdiv{Department of Applied Mathematics and Statistics}, \orgname{Stony Brook University}, \country{USA}}

\affil[6]{\orgdiv{Stony Brook World Trade Center
Wellness Program}, \orgname{Renaissance
School of Medicine at Stony Brook
University}, \country{USA}}

\affil[7]{\orgdiv{Department of Psychology}, \orgname{University of Texas at Dallas}, \country{USA}}

\affil[8]{\orgdiv{Department of Medicine}, \orgname{Renaissance School of Medicine at
Stony Brook University}, \country{USA}}

\affil[9]{\orgdiv{Department of Psychiatry}, \orgname{Stony Brook University}, \country{USA}}



\abstract{
Large language models (LLMs) are increasingly being used in a zero-shot (generative) fashion to assess mental health conditions, yet we have limited knowledge on what factors affect their accuracy. 
In this study, we \revisionpanos{use} a clinical dataset of natural language narratives and self-reported PTSD severity scores from 1,437 individuals to comprehensively evaluate the performance of 11 state-of-the-art LLMs. 
To understand the factors affecting \revisionpanos{model's assessment} accuracy, we systematically varied (i) contextual knowledge \revisionpanos{prompted to the models} like subscale definitions, distribution summary, and interview questions, and (ii) modeling strategies including zero-shot vs few shot, amount of reasoning effort, model sizes, structured subscales vs direct scalar prediction, output rescaling and nine ensemble methods.  
Our findings indicate that
(a) LLMs are most accurate when provided with detailed construct definitions and context of the narrative\revisionpanos{, even exceeding human raters agreement with self-reported scores};
(b) increased reasoning effort leads to better estimation accuracy; 
(c) performance of open-weight models (Llama, DeepSeek) plateaus beyond 70B parameters while closed-weight (gpt-o3-mini, gpt-5) \revisionpanos{alternatives} improve with newer generations;
and (d) best performance is achieved when ensembling a supervised model with the zero-shot LLMs. 
\revisionpanos{Beyond agreement with self-reports, LLMs' estimates discriminated PTSD severity from depression, anxiety, and alcohol use, and prospectively predicted future mental healthcare expenditure.
Together, these results suggest that contextual knowledge and modeling strategies meaningfully affect accuracy and clinical utility of LLM-based assessments of PTSD severity.}

}




\maketitle


\section{Introduction}

Accurate mental health assessment is crucial for diagnosis and treatment for conditions such as the Post-Traumatic Stress Disorder (PTSD).
However, access to timely and specialized care for such conditions remain limited, with the scarcity of trained clinicians often leading to delays in diagnosis and treatment, resulting in aggravated mental health conditions~\cite{owusu-etal-2025, morland-etal-2020, kuhn-etal-2020}.
Despite having standard, highly-validated self-report instruments, such as the PTSD Checklist (PCL) \cite{blevins2015posttraumatic}, many clinicians rely instead on unstructured interviews during intake and treatment sessions to form qualitative impressions~\cite{aboraya-et-al}.
Such narrative accounts are valuable because they allow patients to describe symptoms in their own words, capturing nuances that structured scales may miss~\cite{crespo2016trauma, schnurr2022assessment}.
However, 
\textit{systematically} quantifying mental health severity from natural language communication has historically been difficult~\cite{sikstrom-etal-precise, aboraya-et-al}.

Recently, language models have demonstrated accuracy that approaches a theoretical upper-bound, suggesting natural language accounts, already valued by patients and therapists, could be the solution to the assessment bottleneck~\cite{kjell2022natural, kjell2024beyond}.
However, building on over a decade of natural language processing (NLP)~\cite{crespo2016trauma,wtc_son_et_al, eichstaedt2028predicting, schnurr2022assessment,varadarajan-etal-2024-alba, mangalik2024robust}, such accurate measurement has mostly been shown over supervised LLMs (either fine-tuning or utilizing supervised ML over LM-based embeddings).
The use of zero-shot LLMs for mental health is still developing with less consistent results~\cite{hua2025scoping, brickman-et-al-llms-psych-assessments,stade-etal-2024, dergaa-chatgpt-2024} including PTSD \cite{tu-etal-2024-automating}.
Motivated by inconsistent evaluation practices in mental-health LLMs, we systematically study the role of model’s input knowledge and its assessment procedure on its assessment performance.

To address these gaps and move beyond feasibility studies~\cite{tu-etal-2024-automating, vganesan-etal-schema-2025} toward principled evaluation, we conduct a systematic study of LLMs for estimating PTSD severity from open-ended responses. 
We vary two orthogonal axes that prior works have rarely examined in a controlled way: (1) \textbf{contextual knowledge} provided to the model --- symptom definitions and validated scale items, elicitation context (participant criteria and interview prompts), and task-level priors about score distributions and (2) \textbf{modeling strategy} --— model choice (size and “reasoning” capabilities), task formulation (zero- versus few-shot; direct scalar versus subscales predictions), and post-processing (calibration via predictive redistribution and ensembling). 
This study design responds to calls for standardized~\cite{hua2025scoping} and rigorous evaluation of mental-health LLMs, where existing studies are often vignette-based, and lack consistent design of prompts and modeling choices. 
We aim to understand the knowledge and process contents in prompts that contributes for reliable severity estimation. 

Prior NLP work on PTSD and mental health often draws on social-media self-disclosures~\cite{coppersmith-et-al-2014, coppersmith-natural-2018, coppersmith-etal-2014-quantifying} or in-lab tasks~\cite{burdisso-etal-2024-daic}. 
These settings are useful for prototyping but typically suffer selection, social desirability, demographic and platform biases~\cite{shah-etal-2020-predictive, jaidka-2018-fbvstw, salecha-large-2024, jaidka2022cross, ahsan2025elucidating}, limiting clinical utility. 
Existing studies also frame PTSD as a binary diagnosis rather than a continuous severity construct~\cite{coppersmith-et-al-2014, tu-etal-2024-automating}. 
However in clinical care, symptom severity is monitored on a continuum with instruments such as the PCL-5, and recommended cut points vary by population and purpose~\cite{marx2022reliable}.
Together, these choices reduce clinical relevance and make it difficult to judge whether general-purpose LLMs actually outperform the current go-to approach --- supervised language models trained and evaluated on clinically grounded data with validated scales. 
In contrast, we ground assessment in the PTSD Checklist, a widely used DSM-5–aligned continuous severity measure, and use semi-structured, open-ended patient language rather than social media posts. 
Evaluations of general-purpose LLMs have often appeared inconsistent~\cite{ghazarian2024assessment, askari-etal-2025-assessing}. Much of this variability can be attributed to differences in the contextual information provided to the model — what background knowledge is provided~\cite{v-ganesan-etal-2023-systematic}, how it is framed~\cite{ceballos-arroyo-etal-2024-open, sun2024evaluating}, and where it appears~\cite{liu-etal-2024-lost}.
Broad NLP evidence shows that LLMs' performance shifts with the content and placement of information in long inputs, as well as with how domain knowledge is structured.
Surveys of prompting and in-context learning similarly emphasize that supplying task-relevant knowledge is often decisive~\cite{dong-etal-2024-survey}. 
Yet, in clinical mental-health settings, the systematic study of context provision, e.g., symptom definitions, elicitation details, and task-level priors, has been limited compared to ad-hoc prototypes. 
Our design makes this issue explicit by varying what the model knows (contextual knowledge) and how it is used (modeling strategy). 
In line with emerging work on knowledge-grounding (e.g., types of embedded definitions), this structure yields actionable guidance for safer adoption.

\revisionpanos{
Establishing a language-based assessment as a valid clinical instrument requires evaluating its construct validity~\citep{rust2021modern}. 
Typically, the focus is on (a) \textit{convergent validity}, i.e., agreement with the intended construct, while other types of validty are occassionally considered: (b) \textit{discriminant validity}, i.e., specificity to the construct rather than to general distress, and finally, (c) \textit{criterion validity}, i.e, association with external outcomes that the scale was not designed to capture explicitly. 
This perspective has guided the validation of language-based assessments of constructs such as personality and well-being~\cite{park2015automatic, kjell2024beyond}, yet remains relatively underexplored for zero-shot LLMs in mental health, where evaluation has largely been confined to accuracy against a single reference (often self-reported scores). 
We therefore examine not only how accurately LLMs (a) converge with standard PTSD severity scales, but also their (b) specificity and discriminant validity by contrasting associations with PTSD against general negative affect and broader psychopathology, as well as (c) criterion (or external) validity for predicting mental health care expenditure.
}

In this work, we 
(1) present a comprehensive evaluation of eleven state-of-the-art LLMs on 1,437 open-ended clinical speech samples linked to PCL-5 severity; 
(2) systematically study factors that isolate the influence of contextual knowledge (symptom definitions, elicitation context, task-level priors) and modeling strategy (model choice, zero- vs. few-shot, direct construct measure vs. tallying measures of its elements, post-processing via calibration and ensembling) on performance; 
(3) provide a head-to-head comparison with supervised baselines, showing that well-contextualized configurations — especially when ensembled with a supervised model can outperform alternatives, whereas naïve, context-light usage of LLMs can reduce accuracy by up to~40\%; \revisionpanos{and (4) move beyond predictive accuracy to assess the construct validity of LLM-based estimates, evaluating their specificity to PTSD relative to general negative affect and broader psychopathology, and their external validity through association with future mental healthcare expenditure.}






\section{Results}
\label{sec:results}

We evaluated the capabilities of a series of LLMs to estimate PTSD symptom severity from self-recorded clinical interviews under both zero- and few-shot scenarios. 
Self-reported PCL scores were available for all participants, enabling direct comparison with model predictions. As a baseline, we included a RoBERTa-based \cite{liu-etal-2019} regression model~\cite{kjell-et-al}. Additionally, two expert human raters estimated the PTSD severity of patients in a subset of 187 interviews. The LLM estimates were compared against the self-reported scores on the full dataset. The best performing ones were also compared against the human expert annotations in the manually annotated subset.

\paragraph{Model Scale and Prompting Strategy: LLaMA-3.1 Instruct performs well, few-shot prompting and scaling beyond 70B offer limited gains}

\begin{figure}[t]
\centering

\begin{subfigure}[t]{\linewidth}
\centering
\includegraphics[
    width=.8\linewidth,
    height=6cm
]{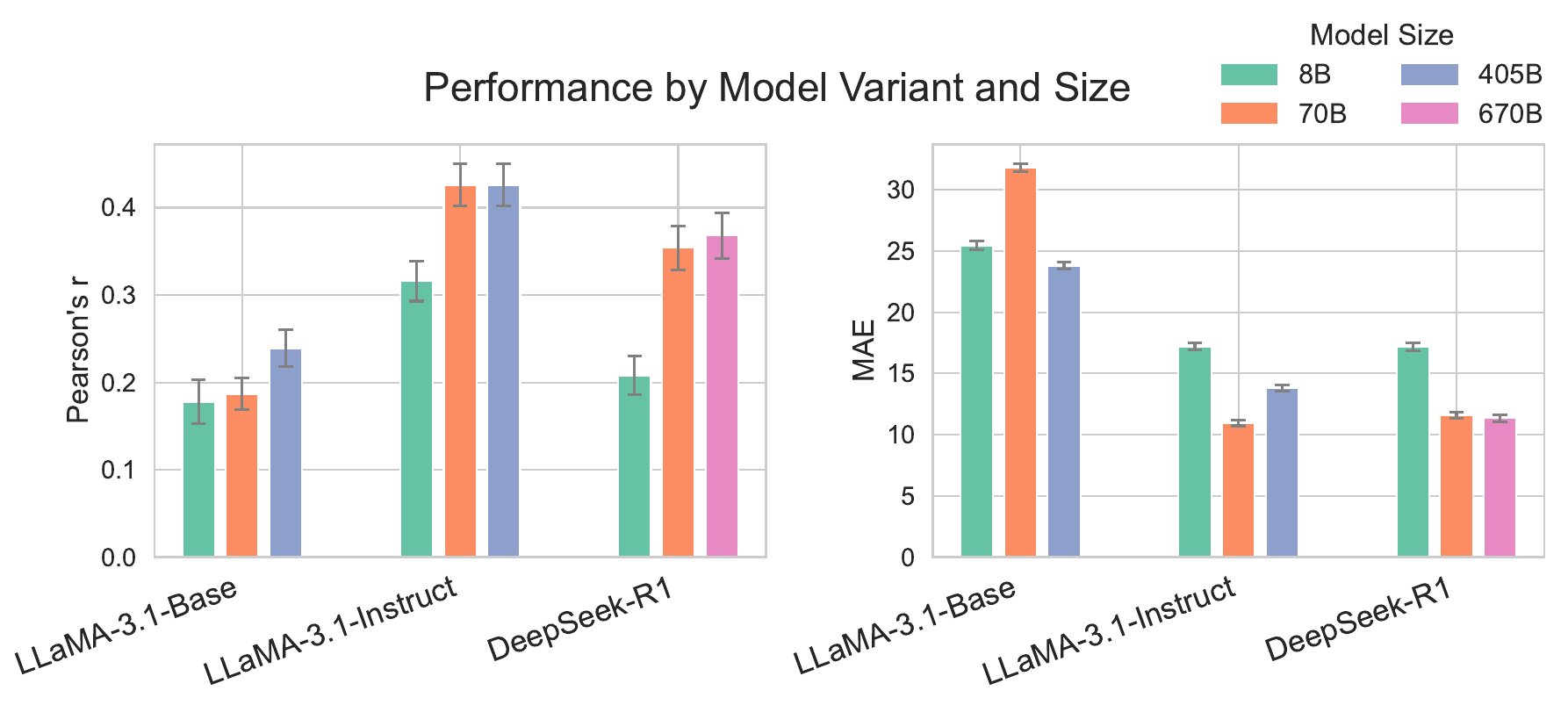}
\caption{Performance scaling for open-weight models.}
\label{fig:model-scaling-sub}
\end{subfigure}

\vspace{0.5em}

\begin{subfigure}[t]{\linewidth}
\centering
\small

\begin{tabular}{lcccc}
\toprule

\multirow{2}{*}{\textbf{Model}}
&
\multicolumn{2}{c}{\textbf{Pearson}}
&
\multicolumn{2}{c}{\textbf{MAE}}
\\

\cmidrule(lr){2-3}
\cmidrule(lr){4-5}

&
0-shot &
Few-shot &
0-shot &
Few-shot
\\

\midrule

Kjell et al. &
\multicolumn{2}{c}{0.421} &
\multicolumn{2}{c}{8.01}
\\

LLaMA-3.1 Base &
.147 & .346 &
31.79 & 15.49
\\

LLaMA-3.1 Inst. &
.426 & .430 &
10.95 & 15.93
\\

gpt-oss &
.366 & .310 &
10.03 & 11.13
\\

LLaMA-405B &
.426 & .361 &
13.80 & 14.69
\\

DeepSeek-R1 &
.368 & .319 &
11.45 & 19.58
\\

4o-mini &
.331 & .301 &
13.66 & 22.75
\\

o3-mini &
.383 & .350 &
8.81 & 10.27
\\

gpt-5 &
\textbf{.441} &
\textbf{.475*} &
\textbf{7.89} &
\textbf{7.95}
\\

\bottomrule
\end{tabular}

\caption{Performance across larger LLMs.}
\label{fig:larger-llm-table}
\end{subfigure}

\caption{
\textbf{Model scaling and performance across larger LLMs.}
\textbf{(a)} Scaling open-weight models beyond 70B parameters did not consistently improve PTSD severity estimation performance, with gains plateauing or degrading for larger variants. 
\textbf{(b)} Comparison across open- and closed-weight LLMs under zero- and few-shot prompting. The strongest-performing open-weight models (LLaMA-3.1-Instruct-70B and DeepSeek-R1-670B) were carried forward together with frontier closed-weight models for subsequent experiments.
}

\label{fig:model_scaling_and_llms}

\end{figure}

We assessed a wide set of high-capacity LLMs, both open-source and proprietary, to examine how PTSD severity estimation performance varies with model size and prompting strategy. LLaMA-3.1-Instruct-70B achieved strong performance, yielding high correlations with self-reported PCL scores. Scaling beyond 70B did not provide consistent gains: larger open models such as DeepSeek-R1 (670B) and proprietary systems with undisclosed sizes (e.g., OpenAI’s 4o-mini and o3-mini) did not surpass the 70B LLaMA-3.1-Instruct model. As illustrated in Figure~\ref{fig:model_scaling_and_llms}a, Pearson correlations plateau past the 70B scale for both LLaMA and DeepSeek variants, with MAE showing similar but less stable trends.  The only model that performed more accurately than the 70B LLaMA variant in both metrics is OpenAI's GPT-5. Moreover, few-shot prompting did not yield consistent improvements and, in some cases, resulted in decreased performance. This was observed, for example, in the cases of DeepSeek-R1 and LLaMA-3.1-Instruct-405B. As a basis of comparison, we included a regression model trained on RoBERTa embeddings, introduced by Kjell et al. \cite{kjell-et-al}. All evaluated models and configurations are presented in Figure \ref{fig:model_scaling_and_llms}b. A comprehensive evaluation across all models, sizes and evaluations is presented in Table S7 (Supplementary Material).


\paragraph{Thinking Step by Step: Inconsistent Performance in Chain-of-Thought Style Reasoning}

We evaluated the impact of Think-Step-By-Step (TSBS) prompting \cite{kojima2022large}, a variant of chain-of-thought (CoT) prompting \cite{wei-etal-2022} on PTSD severity estimation. TSBS encourages the model to verbalize intermediate reasoning steps prior to producing its assessment. As shown in Table~\ref{tab:tsbs-impact}, its effects were inconsistent: TSBS produced small gains for the 8B and 405B LLaMA-3.1 models, but slightly degraded performance for the strong performing 70B model, as well as for OpenAI’s 4o-mini. These mixed results suggest that CoT-style prompting does not reliably enhance performance in this clinical regression task and may even introduce noise for models that already possess strong implicit reasoning capabilities.

\begin{table}[h]
\centering
\caption{ \textbf{Effect of \textit{Think-Step-By-Step} prompting}: Performance comparison between standard and TSBS prompting for 0-shot PCL score estimation. We report Pearson correlation (↑) and MAE (↓) against the self-reported PCL scores. The star symbol (*) denotes configurations whose performance difference against their respective non-TSBS configuration is statistically significant following a non-parametric bootstrapped resampling test.}
\label{tab:tsbs-impact}
\begin{tabular}{lcccc}
\toprule
\textbf{Model} & \textbf{Size} & \textbf{TSBS} & \textbf{Pearson ↑} & \textbf{MAE ↓} \\
\midrule
LLaMA-3.1-Instruct & 8B & \xmark & .316 & 17.21 \\
LLaMA-3.1-Instruct & 8B & \cmark & \textbf{.322} & \textbf{17.18} \\
\midrule
LLaMA-3.1-Instruct & 70B & \xmark & \textbf{.426} & \textbf{10.95} \\
LLaMA-3.1-Instruct & 70B & \cmark & .412 & 11.59 \\
\midrule
gpt-oss & 120B & \xmark & \textbf{.366} & 10.03 \\
gpt-oss & 120B & \cmark & .332 & \textbf{9.92} \\
\midrule
LLaMA-3.1-Instruct & 405B & \xmark & .426 & 13.80 \\
LLaMA-3.1-Instruct & 405B & \cmark & \textbf{.429} & \textbf{11.35}* \\
\midrule
4o-mini & Undisclosed & \xmark & \textbf{.331} & \textbf{13.66}* \\
4o-mini & Undisclosed & \cmark & .311 & 14.15 \\
\bottomrule
\end{tabular}
\end{table}

\paragraph{The effect of reasoning effort: Higher reasoning effort leads to improved model performance}

Some recent LLMs released by OpenAI provide users with the ability to regulate the model’s \textit{reasoning effort} when replying to a given query.
We experimented with all three reasoning effort levels (low, medium and high) of 4o-mini and present results in Table~\ref{tab:reasoning-effort}.
Notably, reasoning effort has a measurable effect on model performance. While performance on the Pearson correlation metric remains relatively stable between the \textit{low} and \textit{medium} settings, substantial improvements are observed in MAE, suggesting that increased reasoning effort better calibrates score magnitudes.
At the \textit{high} reasoning level, we observe significant gains in both metrics ($r = .422$ and MAE $= 8.23$ compared to $.388$ and $9.56$ respectively in the \textit{low} effort level), indicating that increased reasoning effort improves the model’s ability to generate accurate assessments. 
Notably, the average number of reasoning tokens grows nearly fivefold across effort levels, from $453$ to $2237$, suggesting a steep increase in the model’s internal reasoning effort.

\setlength{\tabcolsep}{9pt}
\begin{table}[ht]
\centering
\caption{\textbf{Effect of reasoning effort on PCL score estimation.}
We compare the performance of \texttt{o3-mini} and \texttt{gpt-5} across all reasoning effort levels, while also reporting the average number of \textit{reasoning tokens} per response. Increasing reasoning effort improves MAE consistently and leads to a notable gain in Pearson correlation at the highest level. The star symbol (*) denotes configurations whose performance difference against the low reasoning effort configuration is statistically significant following a non-parametric bootstrapped resampling test.}
\label{tab:reasoning-effort}
\begin{tabular}{c | c | c | S[table-format=1.3] S[table-format=2.2]}
\toprule
\textbf{Model} & \textbf{Reasoning Effort} & \textbf{Avg. \# Reasoning Tokens} & \textbf{Pearson ↑} & \textbf{MAE ↓} \\
\midrule
\multirow{3}{*}{\texttt{o3-mini}} 
  & Low    & 453.4  & 0.388 & 9.56 \\
  & Medium & 1219.1 & 0.383 & 8.81* \\
  & High   & 2237.9 & \textbf{0.422}* & \textbf{8.23}* \\
\addlinespace
\multirow{3}{*}{\texttt{gpt-5}} 
  & Low    & 512.2  & .438 & 8.18 \\
  & Medium & 1109.9  & .441 & 7.89 \\
  & High   & 1896.4  & \textbf{0.451} & \textbf{7.72} \\
\bottomrule
\end{tabular}
\end{table}

\paragraph{The effect of Predictive Redistribution: Post-hoc, Distribution-Aware Adjustments Improve Regression Accuracy}

A common limitation in regression with LLMs is the mismatch in variance and distribution between predicted and target values \cite{geman-etal-1992}. To address this issue, we experimented with a distribution-aware adjustment method known as predictive redistribution \cite{giorgi2022}, which we explain in Section \ref{subsec:analysis}. Table~\ref{tab:redistribution} presents the effect of predictive redistribution across our strongest models and prompting configurations. The most substantial gains appear in MAE, with post-redistribution errors consistently lower in both zero- and few-shot settings. For example, LLaMA-3.1-Instruct (70B) improves from an MAE of $10.95$ to $8.25$ in the zero-shot condition, achieving the best overall result. In contrast, Pearson correlation shows only modest gains in most cases, suggesting that redistribution primarily improves the calibration of absolute predictions rather than their relative ranking. Nonetheless, the highest correlation score ($r = .454$) is achieved in the 3-shot setting with LLaMA-3.1-Instruct after applying redistribution. 


\setlength{\tabcolsep}{3pt}
\begin{table}[ht]
  \centering
  \caption{\textbf{Evaluation of predictive redistribution approach.} We report results for 70B-scale and proprietary models across both zero- and few-shot settings. Predictive redistribution yields substantial improvements in MAE and modest gains in Pearson correlation. The best overall performance is achieved by LLaMA-3.1-Instruct (70B) with redistribution in the 3-shot condition. The star symbol (*) denotes configurations whose performance post predictive redistribution is statistically significantly better following a non-parametric bootstrapped resampling test.}
    \label{tab:redistribution}
    \begin{tabular}{llcccc|cccc}
    \toprule
    \multirow{2}{*}{\textbf{Model Variant}} &
    \multirow{2}{*}{\textbf{Size}} &
    \multicolumn{4}{c|}{\textbf{Pearson (↑)}} &
    \multicolumn{4}{c}{\textbf{MAE (↓)}} \\
    \cmidrule(lr){3-6} \cmidrule(lr){7-10}
    & & 0-shot & 3-shot & 0-shot & 3-shot & 0-shot & 3-shot & 0-shot & 3-shot \\
    & & \multicolumn{2}{c}{Original} & \multicolumn{2}{c|}{Redistributed} & \multicolumn{2}{c}{Original} & \multicolumn{2}{c}{Redistributed} \\
    \midrule
    LLaMA-3.1-Base                  & 70B & .147 & .346 & \hspace{0.14cm}.191* & .345 & 31.79 & 15.49 & \hspace{0.14cm}9.42* & \hspace{0.14cm}9.51* \\
    LLaMA-3.1-Instruct              & 70B & .426 & .430 & \hspace{0.14cm}.434* & \hspace{0.14cm}\textbf{.454}* & 10.95 & 15.93 & \hspace{0.14cm}8.25* & \hspace{0.14cm}\textbf{8.22}* \\
    LLaMA-3.1-Instruct w/ TSBS       & 70B & .412 & .396 & .421 & .401 & 11.59 & 12.58 & \hspace{0.14cm}8.55* & \hspace{0.14cm}8.68* \\
    4o-mini                         & N/A  & .331 & .301 & .336 & .310 & 13.66 & 22.75 & \hspace{0.14cm}9.23* & \hspace{0.14cm}9.37* \\
4o-mini w/ TSBS                  & N/A  & .311 & .318 & .314 & .325 & 14.15 & 18.94 & \hspace{0.14cm}9.33* & \hspace{0.14cm}9.37* \\
o3-mini                         & N/A  & .383 & .350 & .386 & .351 & 8.81  & 10.27 & 8.85 & \hspace{0.14cm}9.04* \\
gpt-5                         & N/A & .441 & .438 & .475 & .476 & 7.89 & 7.95 & 8.49 & \hspace{0.14cm}8.17 \\
    \bottomrule
    \end{tabular}
    \end{table}

\paragraph{Subscale-based vs. Direct Construct Predictions: The Role of Contextual Cues and Scoring Format in PTSD Severity Estimation}

\setlength{\tabcolsep}{5pt}
\begin{table}[h!]
  \centering
  \caption{\textbf{Performance of \textsc{LLaMA-3.1-70B-Instruct} across contextual cue configurations for PTSD severity estimation.} We report both original predictions and redistributed scores for Pearson correlation (↑) and MAE (↓), under zero-shot and 3-shot prompting. The top section reflects subscale-based predictions, and the bottom section shows direct predictions. The star symbol (*) denotes configurations statistically significantly better than the \textit{w/o contextual cues} counterpart (bootstrapped resampling test).}
  \label{tab:contextual-cues-merged}
  \begin{tabular}{l l | cc | cc}
    \toprule
    \textbf{Contextual Cue Configuration} 
      & \textbf{Prompt} 
      & \multicolumn{2}{c|}{\textbf{Pearson ↑}} 
      & \multicolumn{2}{c}{\textbf{MAE ↓}} \\
    \cmidrule(lr){3-4} \cmidrule(lr){5-6}
      & & Orig. & Redistr. & Orig. & Redistr. \\
    \midrule
    \multicolumn{6}{l}{\textbf{Subscale-based Prediction}} \\
    \addlinespace
    \multirow{2}{*}{w/o contextual cues} 
      & 0‐shot & .407 & .417 & 13.91 & 8.50 \\
      & 3‐shot & .377 & .388 & 15.03 & 9.18 \\
    \addlinespace
    \multirow{2}{*}{w/ evidence}  
      & 0‐shot & .412 & .414 & 13.37 & 9.84 \\
      & 3‐shot & \hspace{0.14cm}.414* & \hspace{0.14cm}.415* & \hspace{0.14cm}12.95* & 9.32 \\
    \addlinespace
    \multirow{2}{*}{w/ subscales} 
      & 0‐shot & .426 & \hspace{0.14cm}.434* & 10.95 & 8.25 \\
      & 3‐shot & \hspace{0.14cm}.430* & \hspace{0.14cm}\textbf{.455}* & 15.93 & \textbf{8.22} \\
    \addlinespace
    \multirow{2}{*}{w/ questions} 
      & 0‐shot & \hspace{0.14cm}.417* & \hspace{0.14cm}.419* & 16.77 & 11.22 \\
      & 3‐shot & .367 & .376 & 17.04 & 11.79 \\
    \addlinespace
    \multirow{2}{*}{w/ study context} 
      & 0‐shot & .402 & .411 & \hspace{0.14cm}12.10* & 8.65 \\
      & 3‐shot & .383 & .400 & \hspace{0.14cm}14.01* & 8.90 \\
    \addlinespace
    \multirow{2}{*}{w/ items} 
      & 0‐shot & \hspace{0.14cm}.416* & \hspace{0.14cm}.430* & \hspace{0.14cm}12.00* & 8.40 \\
      & 3‐shot & \hspace{0.14cm}.400* & .406 & \hspace{0.14cm}13.74* & 8.84 \\
    \addlinespace
    \multirow{2}{*}{w/ distributional information} 
      & 0‐shot & .407 & .412 & \hspace{0.14cm}9.62* & 8.61 \\
      & 3‐shot & \hspace{0.14cm}.414* & .422 & \hspace{0.14cm}13.01* & 9.12 \\
    \addlinespace
    \multirow{2}{*}{w/ (evidence, subscales, questions)} 
      & 0‐shot & .392 & .406 & 17.13 & 8.61 \\
      & 3‐shot & .401 & .402 & 12.98 & 9.48 \\
    \midrule
    \multicolumn{6}{l}{\textbf{Direct Prediction}} \\
    \addlinespace
    \multirow{2}{*}{w/o contextual cues}
      & 0‐shot & .443 & .450 & 11.02 & 8.37 \\
      & 3‐shot & .439 & .442 & 12.60 & 8.36 \\
    \addlinespace
    \multirow{2}{*}{w/ evidence}  
      & 0‐shot & \hspace{0.14cm}.467* & .468 & \hspace{0.14cm}10.39* & \hspace{0.14cm}8.08* \\
      & 3‐shot & .463 & .464 & \hspace{0.14cm}11.15* & 8.18 \\
    \addlinespace
    \multirow{2}{*}{w/ subscales} 
      & 0‐shot & .463 & .469 & \hspace{0.14cm}10.03* & 8.18 \\
      & 3‐shot & .436 & .439 & 13.55 & 8.45 \\
    \addlinespace
    \multirow{2}{*}{w/ questions} 
      & 0‐shot & .412 & .415 & 12.34 & 8.43 \\
      & 3‐shot & .438 & .441 & \hspace{0.14cm}11.77* & 8.27 \\
    \addlinespace
    \multirow{2}{*}{w/ study context} 
      & 0‐shot & \hspace{0.14cm}.470* & \hspace{0.14cm}.480* & \hspace{0.14cm}9.66* & \hspace{0.14cm}8.05* \\
      & 3‐shot & .404 & .408 & 12.43 & 8.62 \\
    \addlinespace
    \multirow{2}{*}{w/ items} 
      & 0‐shot & .452 & .456 & \hspace{0.14cm}10.35* & 8.33 \\
      & 3‐shot & .440 & .443 & \hspace{0.14cm}11.03* & 8.27 \\
    \addlinespace
    \multirow{2}{*}{w/ distributional information} 
      & 0‐shot & .452 & .454 & \hspace{0.14cm}\textbf{7.80}* & \hspace{0.14cm}8.08* \\
      & 3‐shot & \hspace{0.14cm}.478* & \hspace{0.14cm}\textbf{.482}* & \hspace{0.14cm}9.18* & \hspace{0.14cm}8.03* \\
    \addlinespace
    \multirow{2}{*}{w/ (evidence, subscales, questions)} 
      & 0‐shot & .413 & .417 & 12.29 & 8.42 \\
      & 3‐shot & .450 & .453 & \hspace{0.14cm}11.61* & 8.18 \\
    \bottomrule
  \end{tabular}
\end{table}

We evaluated two formulations for PTSD severity estimation: \textit{subscale-based prediction}, in which the model predicts the four PCL subscales (Re-experiencing, Avoidance, Dysphoria, Hyperarousal) and we aggregate them into a total score, and \textit{direct construct prediction}, in which the model outputs a single severity value. The former mirrors the structure of the PCL and aligns with clinical practice, whereas the latter simplifies the task. All experiments used LLaMA-3.1-Instruct-70B, which offered strong performance at substantially lower computational cost than 405B. We tested a wide range of contextual cue configurations, which we present extensively in Table S8 (Supplementary Material).

Table~\ref{tab:contextual-cues-merged} presents results across contextual cue configurations for both the \textit{subscale-based} and \textit{direct construct prediction} settings. In the subscale-based setting, Pearson correlations range from $.367$ to $.455$, with vast improvements in MAE following predictive redistribution. The best configuration for this setting (\textit{w/ subscales}, 3-shot) achieved $r = .455$, surpassing the regression-based baseline~\cite{kjell-et-al}. Other cues, such as study context, item descriptions, or questionnaire metadata, provided smaller benefits, whereas not providing any contextual cues led to a performance drop.
The direct construct prediction setting, although it deviates from the formal clinical structure of the PCL, yielded stronger performance under multiple configurations. Prompts including study context ($r = .480$) or distributional information ($r = .482$, MAE = $7.80$) led to the best performance, with item-level cues and subscale definitions also performing well. 

\paragraph{Assessing Specificity Beyond General Negative Affect}


\revisionpanos{We estimated associations with PANAS Negative Affect (NA)~\cite{watson1988development} to evaluate whether LLM estimates primarily reflected broad affective distress rather than PTSD-specific signal. 
Because PANAS measures were available only for a subset of participants, these analyses were performed on the common subset with complete PCL, LLM estimate, and PANAS score ($N=705$). 
On this subset, LLM-predicted PTSD severity correlated with Negative Affect ($r=0.384$), but substantially more strongly with self-reported PCL scores ($r=0.634$). 
Self-reported PCL scores themselves were also associated with Negative Affect ($r=0.498$), as expected~\cite{watson2009differentiating, dornbach2020positive}. 
Importantly, the difference between the correlations of LLM estimates with PCL and Negative Affect was statistically significant (Steiger test: $t=8.61$, $p<0.001$). 
Furthermore, after controlling for Negative Affect, LLM estimates remained strongly associated with PTSD severity (partial $r=0.553$, $p<0.001$).}

\revisionpanos{We extended our analysis on whether LLMs rely on mainly reflecting broad affective distress by correlating the LLM estimates with broader measures of psychological distress. 
Across multiple model and prompting configurations, PTSD predictions consistently showed stronger associations with PTSD-related outcomes than with broader psychopathology measures (Table~\ref{tab:ptsd_specificity}). 
GPT-5 predictions correlated strongly with target self-reported PCL scores ($r=0.522$) while showing near-zero association with alcohol-use severity measured by AUDIT ($r=0.005$). 
Similarly, associations with clinical diagnoses from electronic health records were substantially stronger for PTSD than for anxiety or depression (PTSD $AUC=0.753$, anxiety $AUC=0.638$, depression $AUC=0.694$). 
Together, these findings suggest that while LLM estimates capture affective information related to PTSD, they are not reducible to general negative affect alone and retain substantial PTSD-specific signal.}

\begin{table}[h]
\small
\centering
\caption{\textbf{Associations between LLM-estimated PTSD severity and related clinical measures across model configurations.} Pearson correlations ($r$) are reported for associations with self-reported PCL and alcohol-use severity (AUDIT), while area under the ROC curve (AUC) values are reported for predicting clinical certifications of PTSD, anxiety (ANX), and depression (DEP). Across model and prompting configurations, PTSD estimates consistently showed stronger associations with PTSD-related outcomes than with broader psychopathology measures. Best results per task are underlined.}
\label{tab:ptsd_specificity}

\begin{tabular}{lcc|ccc}
\toprule

\multirow{2}{*}{\textbf{Model Configuration}}
&
\multicolumn{2}{c|}{\textbf{Pearson Correlation ($r$)}}
&
\multicolumn{3}{c}{\textbf{AUC}} \\

\cmidrule(lr){2-3}
\cmidrule(lr){4-6}

&
\textbf{PCL}
&
\textbf{AUDIT}
&
\textbf{PTSD}
&
\textbf{ANX}
&
\textbf{DEP}
\\

\midrule

GPT-5 + Subscale Defs &
\underline{0.522} & 0.005 &
\underline{0.753} & 0.638 & 0.694 \\

LLaMA-3.1-70B + Distr. Info &
\underline{0.452} & 0.008 &
\underline{0.738} & 0.615 & 0.674 \\

LLaMA-3.1-70B + Subscale Defs &
\underline{0.463} & 0.008 &
\underline{0.733} & 0.631 & 0.683 \\

LLaMA-3.1-70B + Study Context &
\underline{0.471} & 0.007 &
\underline{0.741} & 0.634 & 0.680 \\

\bottomrule
\end{tabular}

\end{table}

\begin{figure}[h]

\centering

\includegraphics[width=\linewidth, height=9cm]{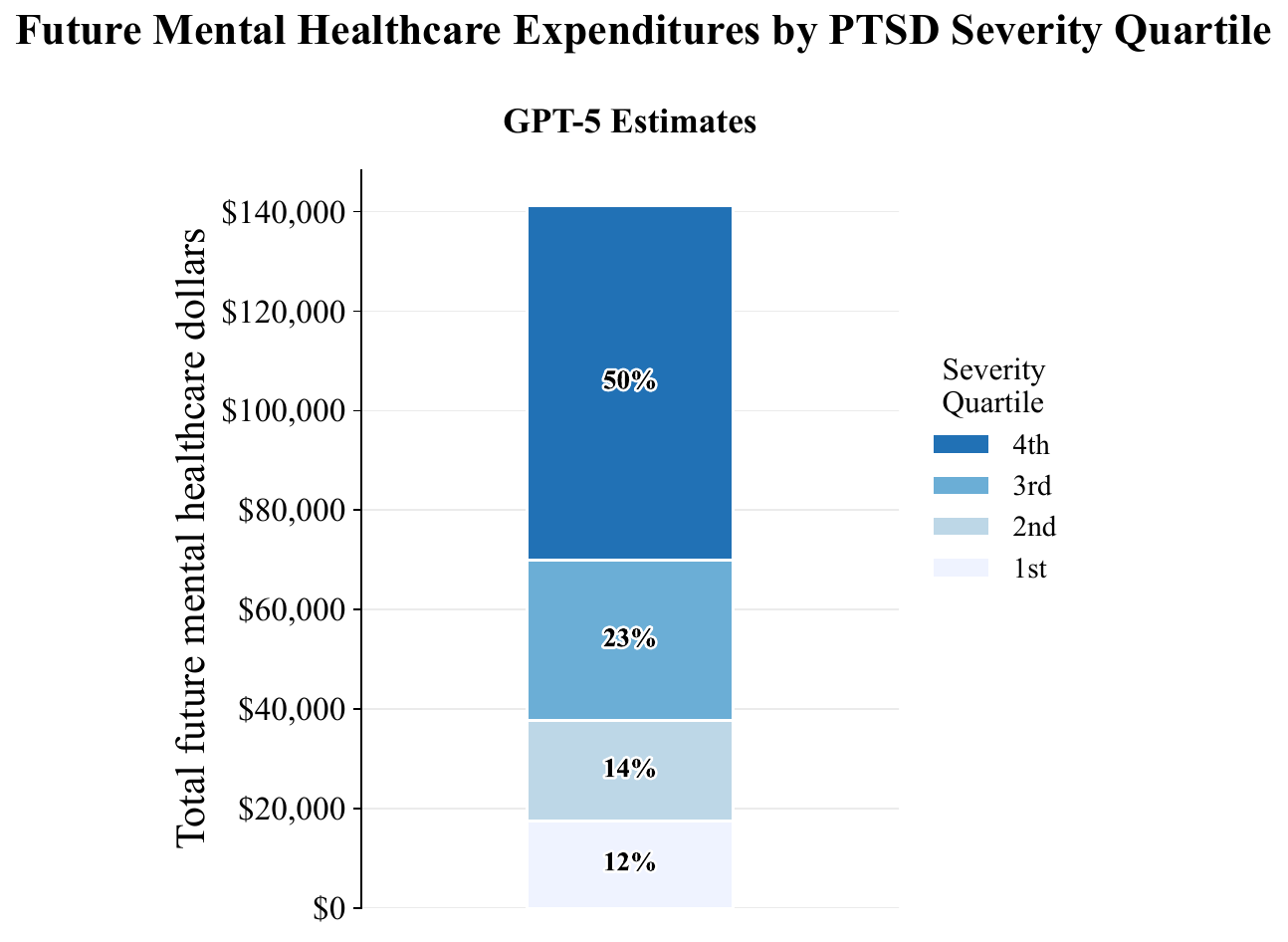}
\caption{\revisionpanos{\textbf{Future mental healthcare expenditures stratified by
GPT-5-estimated PTSD severity.} Participants were divided into quartiles based on GPT-5-estimated PTSD severity. Stacked bars show the proportion of total mental healthcare expenditures incurred at the first subsequent clinical follow-up visit attributable to each severity quartile. Going from lowest estimated severity to highest, each quartile accounted for a progressively larger proportion of future expenditure, where the 20\% of individuals in the highest severity quartile accounted for half the expenditure.}}

\label{fig:future_mental_health_cost}

\end{figure}

\paragraph{Assessing External Validity: Future Mental Healthcare Utilization}
\revisionpanos{
To assess the external validity of language-based PCL estimates, we examined their association with future mental healthcare utilization. 
Specifically, we evaluated whether GPT-5 PTSD severity estimates were associated with total mental healthcare expenditures at the first subsequent clinical follow-up visit, which occurred on average 1.02 years after the interview ($\sigma=0.26$ years). 
GPT-5 estimates were significantly associated with future mental healthcare expenditures ($r=0.268$, $p<0.001$).
Figure~\ref{fig:future_mental_health_cost} provides a complementary stratification analysis. 
Participants in the highest GPT-5-predicted PCL quartile accounted for approximately 50\% of total future mental healthcare expenditures.
These results suggest that language-based PTSD severity estimates capture clinically meaningful information associated with subsequent mental healthcare utilization,
providing evidence of external validity beyond prediction of concurrent symptom severity.}

\paragraph{Incremental Validity Over Self-Reported Scores}
\revisionpanos{
To test whether the language-based estimates carry information beyond self-reports, we examined the incremental validity of the GPT-5 estimates over the self-reported PCL across three outcomes: contemporaneous mental healthcare expenditure, future mental healthcare expenditure at the first subsequent visit, and future PTSD symptom severity. 
For each outcome, we computed the partial correlation between the GPT-5 estimate and the outcome while controlling for the self-reported PCL score. 
Mental healthcare expenditures were log-transformed prior to analysis given their skew. 
Beyond self-report, the GPT-5 estimates remained significantly associated with contemporaneous expenditure (partial $r=0.28$, $p<0.001$) and with future expenditure (partial $r=0.09$, $p<0.001$). 
They also predicted future symptom severity while controlling for a patient's contemporaneous self-reported PCL: the GPT-5 estimate remained associated with their self-reported PCL at the first subsequent visit (partial $r=0.13$, $p<0.001$). 
}

\begin{figure}[!h]
\centering
\includegraphics[width=0.90\linewidth, trim=5 5 5 5, clip]{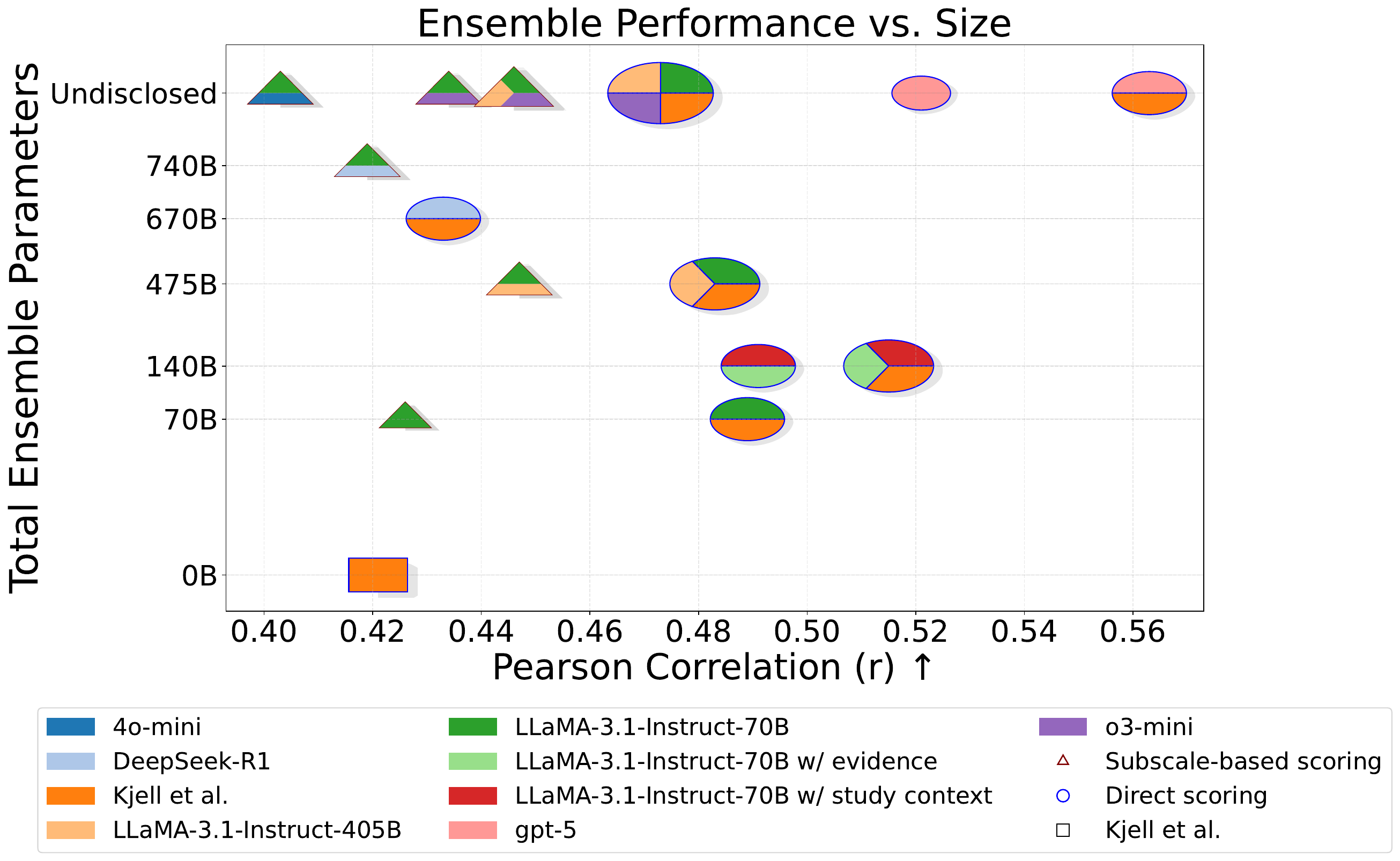}
\caption{
\textbf{Model ensembles performance} on PCL score estimation. Each marker shows an ensemble, positioned by Pearson correlation (x-axis, ↑ better) and MAE (y-axis, ↓ better). Circles denote direct score predictions; triangles denote subscale-based predictions. Marker slices indicate the constituent models (colors per legend), and marker size reflects ensemble size. \revisionpanos{While the much smaller pretrained model of Kjell et al. had relatively lower performance on its own (rectangle on bottom left), when combined with gpt-5 it led to the strongest overall convergence with self-report PCL (circle in top right; $r = 0.57$), suggesting the two models capture different aspects of variance in symptom severity. 
}}
\label{fig:ensembles}
\end{figure}

\paragraph{Model Ensembles: Leveraging Complementary Models for Improved PTSD Severity Estimation}

To examine whether ensembles can further improve PTSD severity estimation, we constructed a wide range of combinations by averaging predictions from complementary systems, including subscale-based models, direct scalar prediction models, and our embedding-based baseline~\cite{kjell-et-al}. Figure~\ref{fig:ensembles} summarizes ensemble performance in terms of Pearson correlation and total number of ensemble parameters. 
Several clear trends emerge. Ensembles built from models prompted to predict severity directly (blue-edged circular markers) generally outperform those relying on subscale prediction (brown-edged triangular markers). The strongest correlation ($r = 0.565$) is achieved by combining GPT-5 (under direct prediction) with the embedding-based baseline. In contrast, ensembles composed of highly similar models show limited improvement. Finally, we observe diminishing returns when ensembling large numbers of models ($n \geq 4$), likely reflecting conflicting predictions that dilute the benefits of diversity. Overall, these findings suggest that ensembles can offer measurable gains in mental health assessment performance~\cite{naderalvojoud-2023, maziarz-etal-2024, wright-etal-2025}.

\paragraph{Comparison with Human Raters}

\begin{figure}[!ht]
    \centering
    \includegraphics[width=0.75\linewidth]{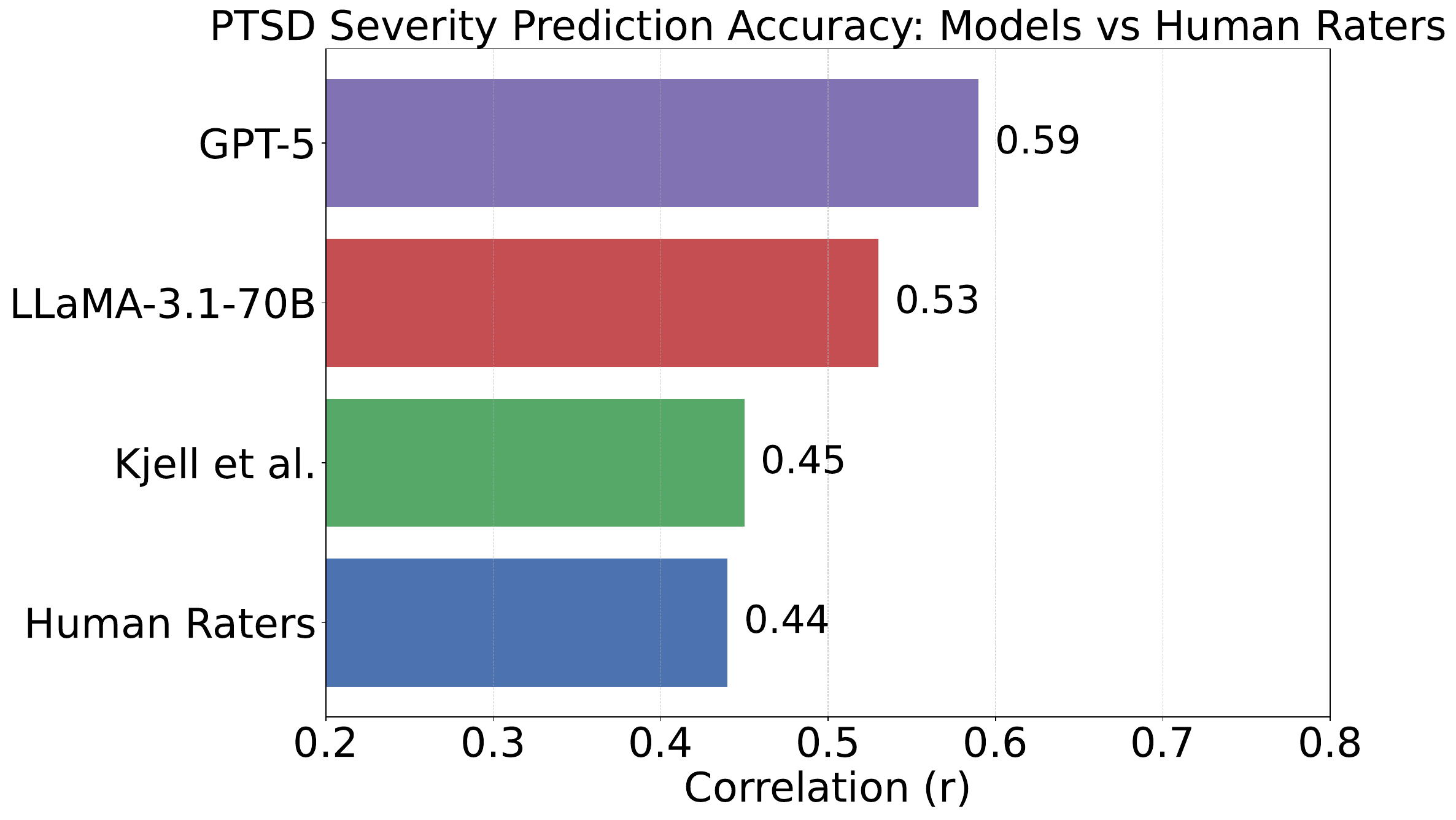}
    \caption{
        Comparison of PTSD severity estimation accuracy across models and human raters.
        Bars show Pearson correlation ($r$) with ground-truth PCL scores on the held-out
        set of 187 interviews. GPT-5 and LLaMA-3.1-70B outperform both human raters and
        the supervised RoBERTa baseline~\cite{kjell-et-al}.
    }
    \label{fig:model_human_comparison}
\end{figure}

To contextualize LLM performance, we additionally collected PTSD severity estimates from two expert human raters. 
The human raters were also informed about the broader study context, including the fact that participants were WTC responders, in order to approximate the background knowledge available in real clinical evaluations. 
Each rater independently reviewed 187 interview transcripts and provided a predicted PTSD severity score. 
We evaluated their predictions against the self-reported PCL scores to establish a human baseline. We present the results in Figure~\ref{fig:model_human_comparison}.
The human raters achieved a correlation of $r = .44$, comparable to our supervised RoBERTa model ($r = .45$) but notably lower than the best-performing LLMs: LLaMA-3.1-70B achieved $r = .53$, and GPT-5 reached $r = .59$. To assess whether these differences were statistically significant, we compared each model’s correlation with that of the human raters using a paired non-parametric bootstrap test. GPT-5 significantly outperformed the human raters ($p = 0.001$), as did LLaMA-3.1-70B ($p = 0.049$), whereas RoBERTa did not differ significantly from human performance ($p = 0.62$). The human raters achieved very high reliability (ICC = $.96$ based on independent double-rating of $187$ transcripts).
These results underscore that frontier LLMs are capable of matching or exceeding the accuracy of trained human raters in zero-shot PTSD severity estimation.


\paragraph{Faithfulness of LLM Rationales}

\revisionpanos{To assess the extent to which LLM PTSD severity estimates are grounded in actual patient
statements versus spurious reasoning, we conducted a human annotation study of
the LLM's reasoning traces. Three trained graduate students independently
annotated a common set of $50$ transcripts at the level of individual claims,
rating each claim on two dimensions: (Q1) whether the claim is grounded in the
patient's language; and (Q2) whether the inference drawn by the LLM is
correct. Full methodological details, rationale segmentation strategy and inter-rater reliability metrics are provided in Section~\ref{sec:methods} and the Appendix.} 

\begin{figure}[t]
    \centering
    \includegraphics[trim=0.05cm 0.2cm 0.3cm 0.15cm, clip, width=\textwidth, height=6cm]{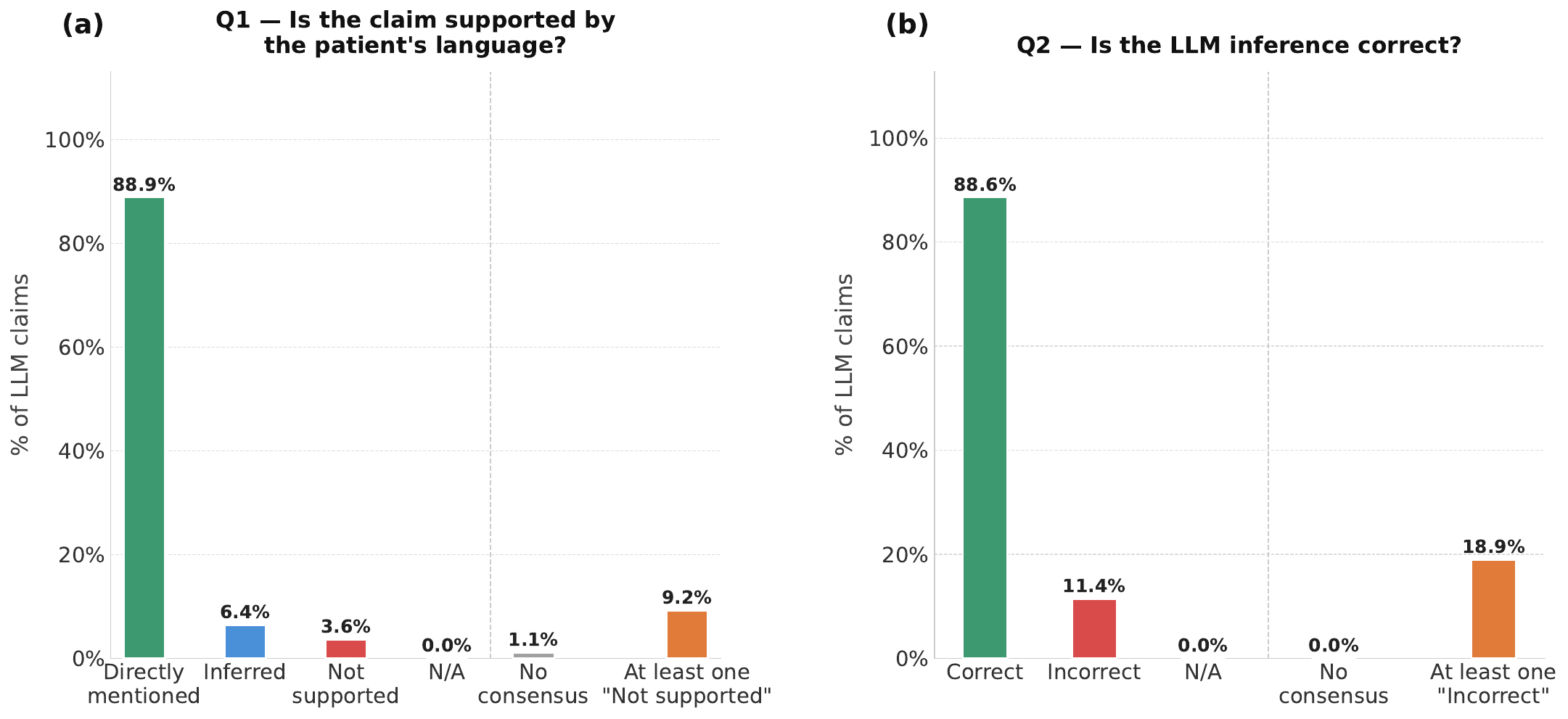}

    \vspace{1.5em}

    \begin{center}
    \begin{tabular}{lcc}
        \toprule
        Dimension & Mean pairwise \% agr. & Krippendorff's $\alpha$ \\
        \midrule
        Q1 (supported vs.\ not supported) & 94.8\% & 0.424 \\
        Q2 (correct vs.\ incorrect)        & 90.9\% & 0.575 \\
        \bottomrule
    \end{tabular}
    \end{center}

    \caption{Annotation study results. \textbf{(a)} Majority-vote consensus
    distributions for Q1: grounding of each claim in the patient's language.
    \textbf{(b)} Majority-vote consensus distributions for Q2: correctness of
    the LLM inference. Bars to the right of the dashed vertical line represent derived
    metrics: \textit{No consensus} indicates claims where no single category
    received a majority vote; \textit{At least one ``Not Supported"/``Incorrect"} indicates claims
    where at least one annotator selected the unfavourable response option,
    irrespective of the majority verdict. The table reports inter-rater
    reliability (via Krippendorff's $\alpha$) for the two annotated questions. Mean pairwise
    percent agreement is computed across the three annotator pairs.}
    \label{fig:hallucination_study_annotations}
\end{figure}

\revisionpanos{
Across the $359$ annotated claims, majority-vote consensus assigned $88.9\%$ as \textit{directly mentioned}, $6.4\%$ as \textit{inferred}, and $3.6\%$ as \textit{not supported}; consensus could not be reached for $1.1\%$ of claims. 
Overall, $95.3\%$ of claims were rated as supported (\textit{directly mentioned} or \textit{inferred}), while at least one annotator flagged $9.2\%$ of claims as \textit{not supported}. 
For Q2, $88.6\%$ of claims were rated as \textit{correct} and $11.4\%$ as \textit{incorrect} under majority vote; at least one annotator flagged $18.9\%$
of claims as \textit{incorrect}. 
These results suggest that the large majority of LLM-generated claims were grounded in the patient's language, though there was a proportion that involved inferences with at least one annotator identifying unsupported or incorrect claim. 
Full distributions are shown in Figure~\ref{fig:hallucination_study_annotations}.}

\section{Discussion}
\label{sec:discussion}


By systematically evaluating eleven state-of-the-art LLMs, we found that LLMs provisioned with specific contextual cues and operated under optimal modeling strategies, can result in higher agreement with self-reported PTSD symptoms than traditional supervised approaches and trained psychologists. 
Although modern LLMs surpassed traditional supervised model, ensembling different LLMs with the supervised model produced the highest agreement with self-reported scores, highlighting the complementary nature of traditional AI methods for language based mental health assessments~\cite{malgaroli-natural-2023}. \revisionpanos{Moreover, LLM estimates showed specificity to the target construct and even incremental validity above it when predicting criteria such as PTSD diagnosis and service utilization from electronic health record.}
These results highlight the potential of carefully configured LLMs to support scalable and timely mental health assessments~\cite{malgaroli2025large, hua-et-al, GalatzerLevy2023TheCO, eberhardt-development-2025}, particularly as pre-screening, monitoring or decision-support tools that can reduce clinician burden~\cite{gaber-etal-2025, bucher-etal-2025}.


Narrative language offers the most natural and comprehensive ways for individuals to express psychological distress. 
Unlike self-report instruments, such as the PTSD Checklist~\cite{blevins2015posttraumatic}, unstructured interviews allow people to describe their symptoms in their own words, capturing nuances in emotion, context, and lived experience that structured instruments may overlook~\cite{ringwald2025uncovering}.
This is why, in practice, many clinicians continue to rely on unstructured interviews to form mental health assessments~\cite{aboraya-et-al}. 
Language based assessments using LLMs can enhance this practice by offering a scalable, automated way to interpret such narrative accounts. 
Automatic LLM-based open-ended language assessments can help reduce delays caused by limited clinician availability~\cite{zhang-etal-2024, katayama-etal-2023} and serve as a bridge for individuals hesitant to seek care due to social stigma~\cite{corrigan2002understanding, habeb-etal-2025}. 

Our findings offer key insights into how contextual knowledge and modeling strategies affect LLM-based PTSD severity estimation. 
First, while increasing model size generally improves accuracy, the gains diminish beyond 70B parameters for open-weight LLMs, indicating a saturation point consistent with trends observed in broader NLP tasks~\cite{mahapatra2025exponential, chowdhery2023palm}, and implying that the scaling laws observed in other domains~\cite{kaplan2020scaling} may not hold for complex clinical estimation tasks like this one. 
Notably, zero-shot prompting often matched or exceeded few-shot performance, aligning with recent studies showing limited marginal gains from few-shot examples in tasks that do not require mathematical or symbolic reasoning~\cite{cheng-etal-2025-revisiting-chain, sprague2025to}. 
Moreover, our results reveal a performance gap between open- and closed-weight models. 
While earlier open-source models performed competitively, GPT-5 significantly outperformed all others, underscoring that proprietary models currently lead on complex clinical inference tasks~\cite{zhang-etal-2024-2, salam-etal-2025}.

\revisionpanos{
Reasoning has increasingly been recognized as a core determinant of LLM performance across complex tasks~\cite{qi-etal-2025-reasoning, kim2024large, kim2025optimizing}, yet our results reveal an important distinction between two fundamentally different forms of reasoning enhancement. Explicit reasoning elicitation through "think-step-by-step" (TSBS) or chain-of-thought (CoT) prompting yielded inconsistent results and sometimes degraded performance, consistent with evidence that CoT provides limited benefit outside mathematical or symbolic tasks~\cite{sprague2025to, cheng-etal-2025-revisiting-chain} and can introduce noise through spurious explanations~\cite{turpin2023language}. However, frontier models explicitly trained for advanced reasoning through reinforcement learning strategies address those limitations~\cite{ouyang2022training, schulman2017proximal, ball2023efficient}. In such models, such as o3-mini and GPT-5, configurable reasoning effort led to a more consistent pattern, with high reasoning effort reliably reducing absolute error and improving correlation, suggesting that the benefit lies not in prompting a model to reason in an explicit manner, but in models trained to systematically reason. Future work could evaluate non-reasoning variants of LLMs against the performance roofline established by their reasoning counterparts, clarifying the marginal contribution of reasoning training and informing cost-efficient deployment decisions. 
}


Beyond model choice, scale and reasoning configurations, we also explored a wide range of task design decisions, particularly the contextual information provided to the model and the structure of the prediction task itself (Table~\ref{tab:contextual-cues-merged}). 
Certain forms of contextual knowledge reliably improved performance over the no-context baseline, such as the inclusion of subscale definitions and PCL item-level descriptions (see Fig. S5 in the Supplementary Material), which appear to help the model anchor its judgments more accurately. 
Other forms of context, such as study-level descriptions, distributional priors, or the full questionnaire items, showed mixed effects: in some cases they yielded modest gains, while in others they had adverse impact, suggesting that not all auxiliary information is beneficial for LLMs in this setting. 

We also compared two task framings: (a) predicting subscale scores individually and (b) directly predicting the overall PCL score. 
Interestingly, unlike the clinician workflow, which is structured around self-reported subscale scores, LLMs achieved stronger performance when asked to predict the total score directly.
\revisionpanos{One possible explanation is that the predicted subscales were substantially less intercorrelated than the corresponding self-reported subscales, reaching an average pairwise $r=0.53$ for our best open-source LLM configuration versus $r=0.71$ for the self-reported subscale scores. This might suggest that decomposing PTSD severity into multiple symptom dimensions may introduce additional prediction noise that accumulates during score aggregation. This result raises broader questions regarding the latent representations learned by LLMs and the clinical constructs they rely upon during prediction. We leave further explanation of such differences in model’s assessment behaviors to future work~\cite{ganesan2026word}.}

In addition, post-hoc redistribution and simple ensembling helped further calibrate predictions, reducing variance and improving alignment with self-reported PCL scores. 
\revisionpanos{
Since the ensemble is comprised of a much smaller fine-tuned roberta-large compared to GPT-5, its computational cost is practically similar to just running GPT-5. 
There would be no change to inference cost, when models are run parallely.
Beyond these practical considerations, our ensemble results carry a conceptual implication. 
A common concern in the field is that large generalist models have rendered smaller, task-specific models obsolete~\cite{sushil2024comparative, ntinopoulos2025large}.
Yet while GPT-5 outperformed a supervised model trained specifically to predict PCL scores, the two were not redundant: the supervised model contributed significant incremental validity. This indicates that the zero-shot approach has weaknesses in psychological assessment that supervised learning can address. Should this pattern hold, ensembling zero-shot and supervised models would enhance predictive power, indicating that natural language contains clinically informative signal that current LLMs alone fail to capture. In this sense, frontier LLMs mark not the end point but the starting point for language-based assessment; complementing their predictions with supervised models may take us closer to precision mental health.
}


\revisionpanos{
The clinical contribution of a language-based assessment lies in anticipating outcomes it was not explicitly designed to capture.
Our external criterion evaluation of models speaks to this directly; 
PTSD severity estimate from a single interview carries information about future mental healthcare expenditure, suggesting it captures prospective functional burden rather than just present symptom level alone~\cite{eichstaedt2028predicting, hur2024language}.
Notably, these estimates showed incremental validity over patients' own self-reports wherein the GPT-5 estimate improved prediction of both current and future mental healthcare costs after accounting for the self-reported PCL, and it forecast future PTSD symptom severity controlling for patients' current symptom levels. 
This indicates that the language-based estimate captures clinically meaningful information not already contained in the self-report questionnaire~\cite{hur2024language}.
Because this language-based signal is cross-construct, prospective, and requires no labeled data from the target population, it points toward a role in triage and resource allocation, helpful in flagging elevated future need before it becomes chronic~\cite{gaber-etal-2025}.
Realizing that role will require validation in treatment-seeking samples and care in treating cost as a proxy for need, but the result establishes that such forward-looking, economically meaningful signal is recoverable from language at intake.
}


This study also comes with limitations. 
The language samples were drawn from a single interview format, so it is unclear how well the models would generalize to other conversational settings~\cite{chandra2025reasoning}. 
The sample consisted exclusively of WTC responders, leaving open questions about generalization to other trauma-exposed groups~\cite{Teferra2025DepressionLLMs, wivine-systematic-2025}. \revisionpanos{However, WTC responders share important demographic, occupational, and communicative characteristics with other populations commonly studied in PTSD research, including military veterans and first responders. Moreover, because the evaluated LLMs operate in a zero-shot setting rather than being fine-tuned on trauma-specific data, their predictions may rely more on general linguistic markers of PTSD symptomatology than on trauma-specific narratives, potentially supporting transferability across trauma contexts.}
Finally, our target objective was the PCL score, and model performance may differ when predicting interviewer-based assessments such as formal diagnoses, despite PCL being the most widely used PTSD severity measure~\cite{pcl17, ptsd-checklist}. 
A further limitation is that our analysis relied solely on textual input. 
Prior work suggests that acoustic and visual modalities can contribute unique information to clinical assessments, including indicators of affect, cognitive load, and physiological arousal~\cite{rao2025whispa, Ringwald2025FacialEmotionPTSD}. 
Incorporating multimodal signals may therefore improve both predictive accuracy and the ecological validity of LLM-based estimators~\cite{sadeghi-etal-2024}. Finally, our study does not address how models internally conceptualize PTSD or arrive at their predictions. 
Recent work highlights the importance of analyzing model reasoning pathways and uncovering latent clinical constructs used by LLMs~\cite{vganesan-etal-schema-2025, yang-etal-2023-towards}, underscoring a need for future work to pair performance evaluations with interpretability analyses.


This study provides a systematic analysis of the factors that shape LLM performance in zero- and few-shot PTSD severity estimation.
Using a clinical dataset of 1,437 interviews, we systematically varied contextual knowledge, task framing, model size, reasoning configuration, output rescaling and ensembling strategies to understand which elements reliably influence accuracy. 
Across these experiments, several patterns emerged consistently: models benefited substantially from certain types of contextual information such as subscale definitions and study context; direct scalar prediction outperformed the clinically common practice of aggregating individual subscale scores; increased reasoning effort did not uniformly improve model performance; model performance plateaued beyond 70B parameters with few-shot prompting offering limited additional gains; and the strongest results arose from ensembling zero-shot LLM predictions with a supervised baseline. 
Language of patients, when assessed through powerful models such as GPT-5, provide incremental value over traditional rating scales in forecasting the traumatic distress levels as well as external outcomes like mental healthcare expenditure.
Together, these findings highlight the the clinical utility of LLMs for mental health assessment, underscoring that effective deployment depends on model choice and also on the careful design of contextual inputs and modeling strategies.

\section{Methods}
\label{sec:methods}

\subsection{Data}
\label{subsec:data}

\paragraph{Collection} All participants take part in the \textit{Stony Brook WTC Health and Wellness Program}~\cite{bromet-etal-2015}. Each participant completed a standardized set of self-recorded questions. Following prior work~\cite{kjell-et-al}, we excluded recordings with fewer than 150 words, yielding a final sample of 1,437 participants. The cohort was predominantly male (92.6\%), with ages ranging from 38 to 90 years (mean = 58.1). Self-recorded interviews averaged 7.5 minutes in duration (SD = 4.1) and ranged between 1.1 and 43.0 minutes. Additional descriptive statistics are provided in Table S9 (Supplementary Material).

\paragraph{Measures} Participants completed an automated clinical interview during a visit, responding aloud to questions displayed on a screen in a private room.  The questions encouraged reflection on past, present, and future experiences, including personal challenges and the impact of the 9/11 events, and instructed participants to speak for at least 60 seconds without reading the prompt aloud. The interview protocol was refined across three iterations to encourage more detailed responses; additional description is provided in \citet{kjell-et-al}. Participants also completed the PTSD Checklist (PCL) \cite{ptsd-checklist}, a 17-item questionnaire designed to assess the severity of PTSD symptoms. Each item is rated on a 5-point Likert scale, from 1 (“not at all severe”) to 5 (“extremely severe”). \revisionpanos{In addition, electronic health record (EHR) data were used for external validation analyses. PTSD-related certifications corresponded to clinician-documented diagnoses of WTC-related PTSD recorded in the medical records at any point since September 11, 2001, based on clinical history and structured diagnostic assessments. Mental healthcare service utilization was operationalized as the total cost of mental health services received by a participant over the previous 12 months, extracted from the WTC electronic health records.} 


\paragraph{Procedure}
Video recordings were collected in a clinical setting at the \textit{Stony Brook WTC Health and Wellness Program}. All participants provided informed consent and were made aware of their right to withdraw at any time. They were guided through the automated interview procedure by a research assistant.

\paragraph{Transcription}

We transcribed each video interview into text using Whisper~\cite{radford2023robust}.
In addition, the transcripts of our study often contain the questions the participants were called to answer, since some of them read them aloud during the interview process. \revisionpanos{As with any ASR-based pipeline, transcription artifacts may propagate into downstream PTSD severity estimation. In particular, omitted words, hallucinated text, or missed negation markers could affect the linguistic evidence available to the LLMs. Future work should further examine the sensitivity of LLM-based clinical assessments to transcript quality.}

\subsection{Experiment Design}
\label{subsec:experiment_design}

\paragraph{Prompt}

The input provided to the model varies depending on the selected prompt configuration, which controls what contextual information is included (see Table \ref{tab:contextual-cues-merged}). Each prompt is composed of a set of core components, present across all configurations, and optional plug-in components that introduce additional information or instructions.

Figures S4 and S6 (Supplementary Material) illustrate the full prompt layout for the subscale-based and direct construct prediction configurations respectively, highlighting the core components (colored boxes) and the three most commonly used optional plug-ins (gray boxes/text spans). The core components include: 
\\(1) the \textbf{Instructions} box (yellow), which defines the overall task for the model.
\\(2) the \textbf{Scoring System} box (green), which outlines a 5-point scale (0--4) with descriptive anchors for each level of symptom severity;  
\\(3) the \textbf{Step-by-Step Procedure} box (blue), which guides the model through the process of symptom detection and score assignment; and  
\\(4) the \textbf{Expected Output} box (red), which specifies the required JSON-like format of the model's output.

In contrast, the optional plug-ins (gray boxes/text spans) act as modular additions that can be toggled on or off depending on the experimental configuration. They are rendered in gray and serve to provide additional context or elicit more grounded reasoning. Additional plug-in elements explored in later configurations are presented in Figure S5 (Supplementary Material). The model performance across all configurations is presented in Table \ref{tab:contextual-cues-merged}. A summary of all plug-in options is provided in Table S8 (Supplementary Material).

\paragraph{Handling of Failure Cases}


Most model outputs followed the expected JSON format, but a small number failed due to malformed or incomplete structure (e.g., altered keys, missing brackets) or by returning free-form text rather than JSON. We addressed these cases using a lightweight validator and a fallback regex-based parser to recover severity scores when possible. Examples that could not be recovered were excluded. Such failures were rare; fewer than 0.5\% across all configurations.

\paragraph{Hyperparameters}

All models were run with temperature set to 0.0 to ensure deterministic outputs and avoid repeated runs, which would be computationally expensive for locally hosted models and costly for closed-source APIs. This allowed performance differences to be attributed to prompt design rather than decoding randomness. Other decoding settings were kept fixed across experiments (top-p = 1.0, frequency penalty = 0.1, presence penalty = 0.0, max tokens = 3000).

\paragraph{Reproducibility and Closed--weight Models} 
\revisionpanos{
Some of the best-performing systems evaluated in this study (e.g., GPT-5, 4o-mini and o3-mini) are proprietary closed-weight models accessed through the Azure OpenAI API. To support reproducibility, we report the exact model identifiers, model release dates, Azure OpenAI API versions and access periods used in our experiments. Specifically, we evaluated gpt-5 (\texttt{2025-08-07}), gpt-4o-mini (\texttt{2024-07-18}), and o3-mini (\texttt{2025-01-31}), all accessed through Azure OpenAI API version \texttt{2024-12-01-preview} during August–-December 2025. Because such systems may evolve through undocumented backend updates, exact model behavior can change over time, posing potential challenges for strict reproducibility in clinical evaluation settings.
}

\paragraph{Human Raters}
To establish a human baseline for comparison, PTSD severity ratings were also collected from two clinical psychology graduate students (“human raters”). Both raters had prior training in diagnostic interviewing and formal instruction in DSM-5 PTSD criteria, and they were supervised throughout the process by a licensed clinical psychologist. Before beginning the main rating task, they completed a structured calibration phase involving 70 interviews for which the true PCL scores were revealed after each rating. This phase helped the raters familiarize themselves with the study population, the structure and content of the transcripts, and the scoring rubric used for PTSD severity. Following calibration, the raters independently evaluated 187 transcripts without access to ground truth. To prevent reliability drift, the raters discussed after every 50 transcripts to resolve discrepancies, without modifying previously assigned scores. This procedure provided tight reliability control across the full rating period.

\paragraph{Methodological Details of Faithfulness Analysis of LLM Rationales}
\revisionpanos{
We developed a web-based annotation tool and recruited three trained graduate
students as annotators to review a stratified sample of $50$ transcripts from
the study dataset. Rationales were generated by LLaMA-3.1-Instruct-70B with
study context (zero-shot). Each rationale was pre-segmented into discrete
claims, yielding $359$ annotated claims across the common set; the segmentation
procedure and exact question wording are provided in the Appendix. Annotators
rated each claim on three dimensions: (Q1) whether the claim is grounded in
the patient's language (\textit{directly mentioned} / \textit{inferred} /
\textit{not supported} / \textit{N/A}); (Q2) whether the inference drawn by
the LLM is correct (\textit{correct} / \textit{incorrect} / \textit{N/A}); and
(Q3), assuming the LLM inference is correct, whether the severity assessment
constitutes an underestimation, overestimation, or appropriate assessment of
the patient's symptom severity. Q3 was piloted but excluded from the final
analysis due to near-chance inter-rater agreement (Krippendorff's $\alpha = 0.038$),
reflecting the inherent difficulty of adjudicating claim-level severity
judgements. For the retained dimensions, inter-rater reliability was assessed
using Krippendorff's $\alpha$ over the $359$ commonly annotated claims:
$\alpha = 0.424$ for Q1 (grouped as \textit{supported} vs.\
\textit{not supported}) and $\alpha = 0.575$ for Q2, indicating moderate to
substantial agreement~\cite{landis1977measurement}.
}

\subsection{Analysis}
\label{subsec:analysis}

\paragraph{Selection of few-shot examples}
To construct the few-shot configurations, we provided the model with three exemplars, each consisting of the speaker's transcript and subscale-level or overall scores. We experimented with two strategies for selecting these examples from the data pool. The first, \textit{range sampling}, ensured that the chosen examples collectively spanned a broad spectrum of PTSD severity scores (e.g., low, moderate, and high). The second, \textit{percentile sampling}, selected instances corresponding to specific quantiles (e.g., 25th, 50th, and 90th percentiles) of the empirical score distribution, ensuring representativeness with respect to the population-level symptom variability. All exemplars were drawn from a held-out set. In preliminary experiments, the \textit{percentile sampling} strategy yielded more consistent improvements in predictive performance compared to range-based selection, thus we proceeded with this for the remainder of the study. A comparison of the two strategies is provided in Table S6 (Supplementary Material). 

\paragraph{Baseline Method}

To better compare the performance of LLM-based scoring approaches, we followed a supervised learning method~\cite{kjell-et-al}, built on top of a frozen RoBERTa backbone and trained using mean squared error (MSE) loss on the self-reported PCL scores. It was evaluated under the same metrics as our experimental setup. While not capable of providing natural language justifications, this baseline serves as a strong text regression reference for direct comparison.

\paragraph{Likert Scale Adjustment}

While the original PTSD Checklist (PCL) employs a 1–5 Likert scale, we modified the scoring prompt to instead use a 0–4 scale. This adjustment was made to better align with the semantics of the task, as assigning a score of ``0'' conveys a more definitive sense of symptom absence (i.e., “not at all severe”) compared to ``1'' \cite{likert-type-scale}.

\paragraph{Predictive Redistribution}
Many regression-based LLMs exhibit the tendency to shrink the variance of their output distributions, often as a side effect of regularization and penalization mechanisms \cite{regression_shrinkage, xie2025variance}. This shrinkage typically pushes predictions toward the mean and produces more normally distributed outputs than the original training targets. While such compression has little effect on correlation-based metrics, it undermines clinical interpretability: for example, predictions may rarely exceed common diagnostic thresholds (e.g., PCL~$\geq$44), limiting usefulness for identifying severe cases. To address this, we applied a \textit{Predictive Redistribution} \cite{giorgi2022}, a two-stage procedure that stabilizes variance before training and restores the target distribution after inference. Specifically, we used the Anscombe transform to rescale scores prior to prediction (Eq.\ref{eq:1}), and then remapped outputs to the empirical mean and variance of PCL scores (Eq.\ref{eq:2}). This adjustment corrects for shrinkage when the underlying score distribution is known, though it is not appropriate in settings lacking prior knowledge about the distribution.

\begin{minipage}{0.45\linewidth}
\begin{equation}
\label{eq:1}
y' = 2 \cdot \sqrt{y + \tfrac{3}{8}} 
\tag{1}
\end{equation}
\end{minipage}
\hfill
\begin{minipage}{0.45\linewidth}
\begin{equation}
\label{eq:2}
y = \left( \frac{y_{\text{pred}} - \mu_{y_{\text{pred}}}}{\sigma_{y_{\text{pred}}}} \right) \cdot \sigma_y + \mu_y
\tag{2}
\end{equation}
\end{minipage}


\paragraph{Reasoning Effort}

OpenAI’s \texttt{o3} and \texttt{GPT-5} model families allow users to specify a \textit{reasoning effort} level when querying the model. This parameter controls how much internal reasoning effort the model performs in order to reach its final answer. Available levels include \textit{low}, \textit{medium}, and \textit{high}, corresponding to increasing amounts of intermediate reasoning. The specific mechanisms through which reasoning effort is regulated remain proprietary, and no detailed public documentation exists about how it influences model behavior.

\paragraph{Statistical Significance Testing}

We assessed whether differences between model performances were statistically significant using non-parametric bootstrap resampling ($n = 1{,}000$). For each metric ($r$, MAE), we repeatedly sampled the evaluation set with replacement and computed the pairwise performance difference between two models. This produced an empirical distribution of differences, from which we derived the 95\% confidence interval (CI). A difference was deemed statistically significant if the CI excluded zero, corresponding to rejecting the null hypothesis at the $p < 0.05$ level.
\section{Declarations}

\paragraph{Data Availability}
The data analysed for this project are not publicly available due to privacy reasons. However, they can be made available from the corresponding author upon request following an anonymization procedure.
\paragraph{Code Availability}
The code for this project will be made available through a GitHub repository. The repository includes scripts for prompting LLMs, processing model outputs, computing evaluation metrics, performing bootstrap significance testing and computing ensembles. Instructions for environment setup will also be provided.

\paragraph{Ethics}

This study was approved by the Stony Brook University Institutional Review Board (approval number IRB2022-00377) and was conducted in accordance with the Declaration of Helsinki. This study does not involve clinical trial.

\paragraph{Funding}

This work was supported in part by grants U01OH012476 and R21OH012614 from CDC/NIOSH to Drs. Andrew Schwartz and Roman Kotov, and by a grant P50 MH 139450 from NIH/NIMH (CREATE: Center for Advancing Therapy with AI) to Dr. Schwartz. 

\paragraph{Acknowledgements}

We express our gratitude to the rescue and recovery workers of the World Trade
Center attacks for their selfless dedication following the WTC attacks and for participating in this
continuous research. Our thanks also extend to the clinical staff of the World Trade Center Medical
Monitoring and Treatment Programs for their unwavering commitment and to the labor and community
organizations for their ongoing support. \revisionpanos{We moreover extend our gratitude to Mia Noel Vasquez, Enisa Adil and Nedret Sen-Cavus, undergraduate/graduate students at University of Texas at Dallas, who served as the human annotators for our study.}

\paragraph{Author Contributions}

AVG, PK, ONEK, RK and HAS designed the research; PK, AVG, and ONEK built the analytic tools; SF and MAC were involved in data collection; Psychology experts WR, ONEK, and RK provided expert feedback throughout the process; CR coordinated human annotation and curated the expert rater data; PK, AVG, ONEK, WR, SF, MAC, DS, CR, BJL, RK and HAS performed the analysis and wrote the manuscript; PK and AVG made all the display items; All authors made the final decision to submit the manuscript.

\paragraph{Competing Interests}

ONEK co-founded and holds shares in a start-up that uses language-based assessments to diagnose mental health problems. The authors report no additional biomedical financial interests or potential conflicts of interest.

\bibliography{sn-bibliography}

\clearpage

\setcounter{figure}{3}
\setcounter{table}{5}

\renewcommand{\thefigure}{S\arabic{figure}}
\renewcommand{\thetable}{S\arabic{table}}

\section*{Supplementary}

\paragraph{Effect of Few-Shot Sampling Strategies on PCL Score Estimation Performance}

This section examines how different few-shot sampling strategies affect model performance in PTSD severity estimation using 70B-scale models. As shown in Table~\ref{tab:sampling_methods}, we compare zero-shot prompting to two few-shot configurations: percentile-based sampling (selecting examples at the 25th, 50th, and 90th percentiles of the PCL score distribution) and range-based sampling (drawing examples from both the center and tail of the distribution). While few-shot prompting generally improves performance for base models like LLaMA-3.1-Base, its effect is less consistent for instruction-tuned models. For instance, LLaMA-3.1-Instruct shows no improvement, or even slight degradation, when few-shot exemplars are added. Across models, percentile sampling tends to slightly outperform range sampling, but gains are modest and vary by architecture. These results highlight that few-shot prompting is not uniformly beneficial and that the choice of example distribution is a sensitive factor in prompt design for clinical prediction tasks.

\begin{table}[ht]
  \centering
  \caption{\textbf{70B models}: We report Pearson correlation (higher is better) and mean absolute error (MAE, lower is better) under three prompting conditions: zero-shot, few-shot with percentile-based sampling (examples at the 25th, 50th, and 90th percentiles of the empirical PCL distribution), and few-shot with stratified sampling (two examples drawn from the central bulk of the distribution and one from its skewed tail).}
  \label{tab:sampling_methods}
  \label{tab:llama70b-original}
  \begin{tabular}{l l l c c}
    \toprule
    \textbf{Model Variant} 
      & \textbf{Size} 
      & \textbf{Prompt} 
      & \textbf{Pearson ↑} 
      & \textbf{MAE ↓} \\
    \midrule
    LLaMA-3.1-Base                  
      & 70B & 0-shot           & 0.147 & 31.79 \\
      &     & 3-shot (range sampling) & 0.345 & 16.73 \\
      &     & 3-shot (percentile sampling)      & \textbf{0.346}   & \underline{15.49}   \\
    \addlinespace
    LLaMA-3.1-Instruct              
      & 70B & 0-shot           & 0.426 & \underline{10.95} \\
      &     & 3-shot (range sampling) & 0.403 & 17.07 \\
      &     & 3-shot (percentile sampling)      & \textbf{0.430}   & 15.93  \\
    \addlinespace
    LLaMA-3.1-Instruct (TSBS)       
      & 70B & 0-shot           & 0.412 & \underline{11.59} \\
      &     & 3-shot (range sampling) & \textbf{0.413} & 13.33 \\
      &     & 3-shot (percentile sampling)      & 0.396   & 12.58   \\
    \addlinespace
    LLaMA-3.1-Instruct w/ distr.\ info  
      & 70B & 0-shot           & 0.407 &  \underline{9.62} \\
      &     & 3-shot (range sampling) & 0.375   &  11.37  \\
      &     & 3-shot (percentile sampling)      & \textbf{0.414}   &  13.01  \\
    \addlinespace
    DeepSeek-Distil-LLaMA           
      & 70B & 0-shot           & 0.354 & \underline{11.65} \\
      &     & 3-shot (range sampling) & \textbf{0.374} & 11.75 \\
      &     & 3-shot (percentile sampling)      & 0.347   &  12.38  \\
    \bottomrule
  \end{tabular}
\end{table}

\paragraph{Comprehensive Performance Comparison Across Model Sizes and Prompting Strategies}

Table~\ref{tab:full-size-comparison} presents a full comparison of model performance across three model size categories (8B, 70B, and larger or undisclosed-scale models), evaluated under consistent prompting protocols. Each model is tested with zero-shot and 3-shot prompting, including standard, instruction-tuned, and chain-of-thought variants. We report both original and redistributed scores for Pearson correlation and MAE. This consolidated result table provides a reference point for understanding how architectural choices, prompting style, and inference strategies jointly shape LLM performance on PTSD severity estimation.

\begin{table}[ht]
\centering
\caption{\textbf{Full model comparison across sizes.} Pearson correlation (↑) and MAE (↓) for PCL score estimation across prompting strategies. We report both original predictions and redistributed scores. Models are grouped by scale.}
\label{tab:full-size-comparison}
\footnotesize
\begin{tabular}{llcccc|cccc}
\toprule
\multirow{2}{*}{\textbf{Model Variant}} &
\multirow{2}{*}{\textbf{Size}} &
\multicolumn{4}{c|}{\textbf{Pearson ↑}} &
\multicolumn{4}{c}{\textbf{MAE ↓}} \\
\cmidrule(lr){3-6} \cmidrule(lr){7-10}
& & 0-shot & 3-shot & 0-shot & 3-shot & 0-shot & 3-shot & 0-shot & 3-shot \\
& & \multicolumn{2}{c}{Original} & \multicolumn{2}{c|}{Redistr.} & \multicolumn{2}{c}{Original} & \multicolumn{2}{c}{Redistr.} \\
\midrule
\multicolumn{10}{l}{\textbf{8B Models}} \\
LLaMA-3.1-Base & 8B & .178 & .251 & .206 & .247 & 25.44 & 26.85 & 10.00 & 10.10 \\
LLaMA-3.1-Instruct & 8B & .316 & .218 & .326 & .217 & 17.21 & 18.72 & 9.30 & 10.04 \\
LLaMA-3.1-Instruct (TSBS) & 8B & .322 & .364 & .328 & \textbf{.365} & 17.18 & 14.19 & 9.28 & \textbf{8.86} \\
\footnotesize DeepSeek-Distil-LLaMA & 8B & .208 & .178 & .221 & .182 & 18.98 & 14.09 & 9.90 & 10.22 \\
\midrule
\multicolumn{10}{l}{\textbf{70B Models}} \\
LLaMA-3.1-Base & 70B & .187 & .345 & .191 & .346 & 31.79 & 16.73 & 9.42 & 9.41 \\
LLaMA-3.1-Instruct & 70B & .426 & .403 & \textbf{.434} & .429 & 10.95 & 17.07 & \textbf{8.25} & 8.49 \\
LLaMA-3.1-Instruct (TSBS) & 70B & .412 & .413 & .421 & .424 & 11.59 & 13.33 & 8.55 & 8.53 \\
DeepSeek-Distil-LLaMA & 70B & .354 & .374 & .364 & .376 & 11.65 & 11.75 & 9.05 & 8.80 \\
\midrule
\multicolumn{10}{l}{\textbf{Larger Model Variants}} \\
LLaMA-3.1-Instruct & 405B & .426 & .361 & .437 & .366 & 13.80 & 14.69 & 8.45 & 8.99 \\
LLaMA-3.1-Instruct (TSBS) & 405B & .429 & .416 & \textbf{.438} & .415 & 11.35 & 12.38 & \textbf{8.27} & 8.51 \\
DeepSeek-R1 & 670B & .368 & .319 & .371 & .322 & 11.45 & 19.58 & 8.93 & 9.99 \\
4o-mini & N/A & .331 & .295 & .336 & .307 & 13.66 & 23.96 & 9.23 & 9.43 \\
4o-mini (TSBS) & N/A & .311 & .294 & .314 & .303 & 14.15 & 20.01 & 9.33 & 9.49 \\
o3-mini & N/A & .383 & .344 & .386 & .349 & 8.81 & 10.74 & 8.85 & 9.11 \\
\bottomrule
\end{tabular}
\end{table}

\paragraph{Optional Plug-In Components for Prompt Customization}

Table~\ref{tab:prompt-plugins} summarizes the optional plug-in components used to construct prompts across experimental configurations. These modular elements were toggled on or off to systematically assess their contribution to model performance in PTSD severity estimation. Each component provides a distinct type of clinical or contextual framing, such as subscale definitions, interview questions, or study background (e.g., post-9/11 context). Other elements, like distributional information or PCL item references, offer structural guidance or calibration cues to shape model outputs. This modular design enabled targeted evaluations of which kinds of information most effectively guide large language models in generating accurate and clinically meaningful predictions.

\begin{table}[h!]
\centering
\caption{Overview of optional plug-in components used in the prompt design. These components can be toggled on or off across configurations to evaluate their effect on model performance.}
\label{tab:prompt-plugins}
\begin{tabular}{@{}p{4.5cm}p{8.2cm}@{}}
\toprule
\textbf{Plug-in Component} & \textbf{Description} \\
\midrule
\textbf{w/ Evidence} & Prompts the model to explain the reasoning behind each subscale severity score prediction by referencing relevant excerpts from the transcript. \\
\midrule
\textbf{w/ Subscale Definitions} & Short descriptions for each of the four PTSD subscales (re-experiencing, avoidance, dysphoria, hyperarousal) to guide the model's interpretation. \\
\midrule
\textbf{w/ Interview Questions} & The set of questions originally posed to participants during self-recorded interviews, aimed at eliciting responses indicative of PTSD symptoms. \\
\midrule
\textbf{w/ Study Context} & An expanded instruction indicating that the interview relates to post-9/11 lived experiences and clarifying the clinical framing. \\
\midrule
\textbf{w/ PCL Items} & The original 17 PTSD Checklist (PCL) items used for clinical scoring \cite{ptsd-checklist}, listed to help the model align its predictions to standard symptom indicators. \\
\midrule
\textbf{w/ Distributional Information} & A summary of expected score distributions to steer the model away from extreme outputs without strong textual evidence. \\
\bottomrule
\end{tabular}
\end{table}


\paragraph{Prompt Architecture for PTSD Severity Estimation Experiments}

Figure~\ref{fig:prompt-structure} presents the full layout of the prompt used for the PTSD severity estimation experiments. The design consists of modular components that were either fixed or toggled depending on the experimental configuration. Core elements shared across all prompts include the task instructions, the structured scoring system, a clearly defined two-step procedure outlining the expected scoring process, as well as the expected output format. Optional plug-in components, shown in gray, include subscale definitions, interview questions and asking for references to parts of the text that motivated the model's decisions. These components were included or excluded in a controlled manner to assess their contribution to model performance.


\begin{center}
\begin{tcolorbox}[
  enhanced,
  breakable,
  sharp corners=south,
  colframe=black,
  boxrule=0.6pt,
  colback=white,
  fonttitle=\bfseries,
  left=0pt, right=0pt, top=0pt, bottom=1pt,
  boxsep=0pt, arc=0pt
]

\begin{tcolorbox}[
  colback=lightyellow,
  colframe=lightyellow,
  boxrule=0pt,
  top=3pt, bottom=2pt, left=6pt, right=6pt,
  boxsep=0pt, enlarge top by=0mm, enlarge bottom by=-1pt,
  fontupper=\small
]
\textbf{Instructions:} Your task is to conduct an in-depth analysis of a provided transcript, with the goal of assessing the severity of PTSD symptoms based on its 4 subscales. The transcript is derived from a patient’s self-recorded answers to a set of questions, designed to assess various aspects of psychological well-being, including PTSD-related symptoms.
\end{tcolorbox}

\begin{tcolorbox}[
  colback=lightgray,
  colframe=lightgray,
  boxrule=0pt,
  top=2pt, bottom=2pt, left=6pt, right=6pt,
  boxsep=0pt, enlarge top by=-1pt, enlarge bottom by=-1pt,
  fontupper=\small,
]
\textbf{Questions Asked:}
\begin{enumerate}[leftmargin=1.5em]
  \item How are you? Can you elaborate?
  \item How's the family? Can you elaborate?
  \item[\hspace{1.5em}] $\cdot$\\[-2.0em]
  \item[\hspace{1.5em}] $\cdot$\\[-2.0em]
  \item[\hspace{1.5em}] $\cdot$\\[-1.5em]
  \setcounter{enumi}{12}
  \item Over the past 5 years what are the three worst things that happened to you and your family? Can you elaborate?
\end{enumerate}
\end{tcolorbox}

\begin{tcolorbox}[
  colback=lightgray,
  colframe=lightgray,
  boxrule=0pt,
  top=2pt, bottom=2pt, left=6pt, right=6pt,
  boxsep=0pt, enlarge top by=-1pt, enlarge bottom by=-1pt,
  fontupper=\small
]
\textbf{PTSD Subscales:}
\begin{itemize}[leftmargin=1.5em]
  \item \textbf{Re-experiencing:} Intrusive thoughts, flashbacks, nightmares, or distress when reminded of the trauma.
  \item \textbf{Avoidance:} Efforts to avoid thoughts, feelings, or external reminders associated with the trauma.
  \item \textbf{Dysphoria:} Persistent negative thoughts, feelings of guilt, emotional numbness, or loss of interest in activities.
  \item \textbf{Hyperarousal:} Increased irritability, hypervigilance, being easily startled, or difficulty concentrating and sleeping.
\end{itemize}
\end{tcolorbox}

\begin{tcolorbox}[
  colback=lightgreen,
  colframe=lightgreen,
  boxrule=0pt,
  top=2pt, bottom=2pt, left=6pt, right=6pt,
  boxsep=0pt, enlarge top by=-1pt, enlarge bottom by=-1pt,
  fontupper=\small
]
\textbf{Scoring System:}
\begin{itemize}[leftmargin=1.5em]
  \item 0 - Not at all severe: The symptom is not present or negligible.
  \item 1 - Low severity: The symptom is present but slightly impacts the individual.
  \item 2 - Moderate severity: The symptom is noticeably present and moderately impacts the individual.
  \item 3 - High severity: The symptom is significantly present and strongly impacts the individual.
  \item 4 - Extreme severity: The symptom is overwhelmingly present and has a profound impact.
\end{itemize}
\end{tcolorbox}

\begin{tcolorbox}[
  colback=lightblue,
  colframe=lightblue,
  boxrule=0pt,
  top=0pt, bottom=0pt, left=6pt, right=6pt,
  boxsep=0pt, enlarge top by=-1pt, enlarge bottom by=-1pt,
  fontupper=\small
]
\textbf{Steps:}
\begin{itemize}[leftmargin=1.5em]
  \item \textbf{Step 1:} Detect the PTSD subfactors in the text \colorbox{gray!15}{\textit{and provide an explanation on how each symptom was identified.}} Provide a score between 0 and 4.
  \item \textbf{Step 2:} Present the results as a nested JSON with severity scores \colorbox{gray!15}{\textit{and explanations referencing relevant spans or contextual inferences.}}
\end{itemize}
\end{tcolorbox}

\begin{tcolorbox}[
  colback=lightpurple,
  colframe=lightpurple,
  boxrule=0pt,
  top=3pt, bottom=4pt, left=6pt, right=6pt,
  boxsep=0pt, enlarge top by=1pt, enlarge bottom by=0pt,
  fontupper=\ttfamily\small
]
\textbf{Expected JSON Output:}
\medskip

\{
\hspace*{1.5em}"Re-experiencing": \{\\
\hspace*{3em}\colorbox{gray!15}{"Reason": [\textit{"<Reason(s) from transcript>"}]},\\
\hspace*{3em}"Severity Score": <score>\\
\hspace*{1.5em}\},\\
\hspace*{1.5em}"Avoidance": \{\\
\hspace*{3em}\colorbox{gray!15}{"Reason": [\textit{"<Reason(s) from transcript>"}]},\\
\hspace*{3em}"Severity Score": <score>\\
\hspace*{1.5em}\},\\
\hspace*{1.5em}"Dysphoria": \{\\
\hspace*{3em}\colorbox{gray!15}{"Reason": [\textit{"<Reason(s) from transcript>"}]},\\
\hspace*{3em}"Severity Score": <score>\\
\hspace*{1.5em}\},\\
\hspace*{1.5em}"Hyperarousal": \{\\
\hspace*{3em}\colorbox{gray!15}{"Reason": [\textit{"<Reason(s) from transcript>"}]},\\
\hspace*{3em}"Severity Score": <score>\\
\hspace*{1.5em}\}
\}
\end{tcolorbox}
\begin{tcolorbox}[
  colback=white,
  colframe=white,
  boxrule=0.3pt,
  top=1pt, bottom=1pt, left=6pt, right=6pt,
  boxsep=2pt, enlarge top by=0pt, enlarge bottom by=0pt,
  fontupper=\small
]
\textbf{Text:} \textit{"I’ve been having a rough time lately. Some days are better than others, but overall, I feel like I’m constantly on edge. I still think about what happened all the time, not always directly, but little things set me off. Loud noises, crowded places. I try to keep busy..."}
\end{tcolorbox}

\end{tcolorbox}

\vspace{0.5em}
\captionof{figure}{Full layout of the PTSD subscale evaluation prompt. Colored boxes represent core components shared across all configurations. Gray boxes are optional plug-ins that can be included or excluded to control the amount of context provided.}
\label{fig:prompt-structure}
\end{center}

\paragraph{Additional Plug-In Components for Further Prompt Customization}

Figure~\ref{fig:prompt-variants} illustrates three plug-in prompt components that were selectively incorporated in different experimental conditions. These elements either modify core instructions or introduce additional clinical context to guide the model’s scoring behavior. Panel (a) shows an updated instruction variant that embeds explicit study context, framing the task around participants’ experiences following the 9/11 World Trade Center attacks. Panel (b) introduces the full list of PTSD Checklist (PCL) items. Panel (c) presents distributional guidance about typical PCL score ranges, intended to calibrate model predictions by discouraging extreme scores unless strongly supported by the input. These components were activated in a controlled manner across configurations to investigate whether they yield more accurate severity estimates.

\pagebreak
\begin{center}
\begin{minipage}{\textwidth}

\begin{tcolorbox}[
  enhanced,
  breakable,
  colback=white,
  colframe=black,
  boxrule=0.6pt,
  sharp corners=south,
  left=3pt, right=3pt, top=5pt, bottom=5pt,
  boxsep=0pt
]

\begin{tcolorbox}[
  colback=lightpink,
  colframe=lightpink,
  width=\linewidth,
  boxrule=0pt,
  top=4pt, bottom=4pt, left=6pt, right=6pt,
  boxsep=0pt, fontupper=\small
]
\textbf{(a) Study Context [\textit{Updated Instruction Component}]:}

Your task is to conduct an in-depth analysis of a provided transcript, with the goal of assessing the severity of PTSD symptoms based on its 4 subscales. The text you will analyze is the transcription of the patient’s self-recorded answers to a set of questions. The patients are questioned \underline{about their lives after the World Trade Center disaster that happened on 9/11}. These questions were designed to assess various aspects of psychological well-being, particularly symptoms associated with PTSD. The goal is to assess the overall PTSD severity \underline{related to the World Trade Center disaster}.
\end{tcolorbox}


\begin{tcolorbox}[
  colback=lightcyan,
  colframe=lightcyan,
  width=\linewidth,
  boxrule=0pt,
  top=4pt, bottom=4pt, left=6pt, right=6pt,
  boxsep=0pt, fontupper=\small
]
\textbf{(b) PCL Items [\textit{Additional Component}]:}

The PCL score is calculated based on 17 specific items, each representing a symptom commonly associated with PTSD. The 17 items are:
\begin{enumerate}[leftmargin=1.5em]
  \item Repeated, disturbing memories, thoughts, or images of a stressful experience from the past.
  \item Repeated, disturbing dreams of a stressful experience from the past.
  \item[\hspace{1.5em}] $\cdot$\\[-2.0em]
  \item[\hspace{1.5em}] $\cdot$\\[-2.0em]
  \item[\hspace{1.5em}] $\cdot$\\[-1.5em]
  \setcounter{enumi}{16}
  \item Feeling jumpy or easily startled.
\end{enumerate}
\end{tcolorbox}

\begin{tcolorbox}[
  colback=lightorange,
  colframe=lightorange,
  width=\linewidth,
  boxrule=0pt,
  top=4pt, bottom=4pt, left=6pt, right=6pt,
  boxsep=0pt, fontupper=\small
]
\textbf{(c) Distributional Information [\textit{Additional Component}]:}

According to domain experts, most individuals typically score between 17 and 30. Scores between 30 and 50 occur in fewer cases, while above 50 are relatively rare and typically indicate more severe or clinically significant symptom expression.
You should take this distribution into account when making your severity judgments. Use higher scores only when there is clear textual evidence of substantial psychological distress or impairment.

\end{tcolorbox}

\end{tcolorbox}

\vspace{6pt}
\captionof{figure}{Further plug-in prompt components that were explored in this study. These elements provide additional information and are activated selectively. They can either be additional components (\textit{Additional Component)} or alternative variants of a core component (\textit{Updated Component}).}
\label{fig:prompt-variants}
\end{minipage}
\end{center}

\paragraph{Direct score prediction prompt}

\begin{center}
\begin{tcolorbox}[
  enhanced,
  breakable,
  sharp corners=south,
  colframe=black,
  boxrule=0.6pt,
  colback=white,
  fonttitle=\bfseries,
  left=0pt, right=0pt, top=0pt, bottom=1pt,
  boxsep=0pt, arc=0pt
]

\begin{tcolorbox}[
  colback=lightyellow,
  colframe=lightyellow,
  boxrule=0pt,
  top=3pt, bottom=2pt, left=6pt, right=6pt,
  boxsep=0pt, enlarge top by=0mm, enlarge bottom by=-1pt,
  fontupper=\small
]
\textbf{Instructions:} Your task is to conduct an in-depth analysis of a provided transcript, with the goal of estimating the overall severity of post-traumatic stress disorder (PTSD) symptoms experienced by the individual.
The text you will analyze is a transcription of the patient's spoken answers to a predefined set of self-recorded questions. The patients are questioned about their lives after the World Trade Center disaster that happened on 9/11. These questions were designed to assess various aspects of psychological well-being, particularly symptoms associated with PTSD. The goal is to assess the overall PTSD severity related to the World Trade Center disaster.

\end{tcolorbox}

\begin{tcolorbox}[
  colback=lightgreen,
  colframe=lightgreen,
  boxrule=0pt,
  top=2pt, bottom=2pt, left=6pt, right=6pt,
  boxsep=0pt, enlarge top by=-1pt, enlarge bottom by=-1pt,
  fontupper=\small
]
\textbf{Scoring System:}
Based on the content of the transcript, predict a single scalar PTSD severity score in the range 17 to 85, where:

- 17 represents minimal or no PTSD-related symptoms.
- 85 represents extreme PTSD symptom severity across multiple domains of functioning.

This score should directly estimate the patient’s PCL score, a widely used self-report measure of PTSD symptom severity.
\end{tcolorbox}

\begin{tcolorbox}[
  colback=lightblue,
  colframe=lightblue,
  boxrule=0pt,
  top=0pt, bottom=0pt, left=6pt, right=6pt,
  boxsep=0pt, enlarge top by=-1pt, enlarge bottom by=-1pt,
  fontupper=\small
]
\textbf{Steps:}
Carefully analyze the transcript, considering the emotional tone, content, and any references to trauma-related symptoms or functional impairments. Then assign a single integer score between 17 and 85 that best reflects the overall PTSD severity of the individual related to the World Trade Center disaster.
\end{tcolorbox}

\begin{tcolorbox}[
  colback=lightpurple,
  colframe=lightpurple,
  boxrule=0pt,
  top=3pt, bottom=4pt, left=6pt, right=6pt,
  boxsep=0pt, enlarge top by=1pt, enlarge bottom by=0pt,
  fontupper=\ttfamily\small
]
\textbf{Expected JSON Output:}
Return your answer in the following structured JSON format:

\medskip

\{
"PTSD Severity Score": <score>
\}
\end{tcolorbox}
\begin{tcolorbox}[
  colback=white,
  colframe=white,
  boxrule=0.3pt,
  top=1pt, bottom=1pt, left=6pt, right=6pt,
  boxsep=2pt, enlarge top by=0pt, enlarge bottom by=0pt,
  fontupper=\small
]
\textbf{Text:} \textit{"I’ve been having a rough time lately. Some days are better than others, but overall, I feel like I’m constantly on edge. I still think about what happened all the time, not always directly, but little things set me off. Loud noises, crowded places. I try to keep busy..."}
\end{tcolorbox}

\end{tcolorbox}

\vspace{0.5em}
\captionof{figure}{Full layout of the PTSD direct score prediction prompt, which is simpler compared to the subscales prediction one shown in Figure~\ref{fig:prompt-structure}.}
\label{fig:prompt-direct}
\end{center}

\paragraph{Dataset Descriptive Statistics}
\begin{table}[ht]
\centering
\caption{Descriptive statistics of our dataset, including demographic information and basic audio characteristics of the self-recorded clinical interviews.}
\label{tab:descriptive-stats}
\begin{tabular}{llc}
\toprule
\textbf{Category} & \textbf{Metric} & \textbf{Analysis Set} \\
\midrule
\multirow{3}{*}{\textbf{Demographics}} 
  & Number of Recordings & 1437 \\
  & Age (min / mean / max) & 38 / 58.09 / 90 \\
  & Gender Ratio (F:M)   & 7.4 : 92.6 \\
\midrule
\multirow{2}{*}{\textbf{Audio Stats}} 
  & Avg. Duration (min) & 7.50 ± 4.1 \\
  & Avg. Word Count     & 697.93 ± 530.44 \\
\bottomrule
\end{tabular}
\end{table}

\paragraph{Faithfulness Analysis of LLM Rationales: Annotation Questions}

The annotation tool presented each claim alongside the full interview transcript
and the model's predicted PCL-5 score. Figure~\ref{fig:annotation-questions}
shows the three questions and their response options as displayed to annotators.

\begin{center}
\begin{tcolorbox}[
  enhanced,
  breakable,
  sharp corners=south,
  colframe=black,
  boxrule=0.6pt,
  colback=white,
  fonttitle=\bfseries,
  left=0pt, right=0pt, top=0pt, bottom=1pt,
  boxsep=0pt, arc=0pt
]

\begin{tcolorbox}[
  colback=lightyellow,
  colframe=lightyellow,
  boxrule=0pt,
  top=3pt, bottom=2pt, left=6pt, right=6pt,
  boxsep=0pt, enlarge top by=0mm, enlarge bottom by=-1pt,
  fontupper=\small
]
\textbf{Instructions to annotators:} You will be provided with claims that were used to make
a PTSD severity assessment of a person. These claims are rationales generated
in the context of assessments of PTSD severity, accompanied by a predicted
PCL-5 score. Your task is to evaluate each claim under two aspects:
(a) whether that claim is supported by what is present in the
patient's language; (b) whether the rationale behind the claim supports the assessment made, and (c) assuming that the inferences the LLM made are correct, evaluate the severity assessment made. 
Note: your task is to judge whether the rationale
supports the score, not whether the score itself is correct.
\end{tcolorbox}

\begin{tcolorbox}[
  colback=lightblue,
  colframe=lightblue,
  boxrule=0pt,
  top=2pt, bottom=2pt, left=6pt, right=6pt,
  boxsep=0pt, enlarge top by=-1pt, enlarge bottom by=-1pt,
  fontupper=\small
]
\textbf{Q1: Is the claim supported by the patient's language?}
\begin{itemize}[nosep, leftmargin=1.4em]
  \item \textbf{(a) Yes, directly mentioned}; the transcript directly states
  what the claim describes.
  \item \textbf{(b) Yes, inferred}; it can be reasonably inferred from what
  the patient said.
  \item \textbf{(c) No, it is not}; the information is neither directly
  mentioned nor inferable.
  \item \textbf{(d) N/A}; the claim cannot be evaluated against the
  transcript.
\end{itemize}
\end{tcolorbox}

\begin{tcolorbox}[
  colback=lightgreen,
  colframe=lightgreen,
  boxrule=0pt,
  top=2pt, bottom=2pt, left=6pt, right=6pt,
  boxsep=0pt, enlarge top by=-1pt, enlarge bottom by=-1pt,
  fontupper=\small
]
\textbf{Q2: Evaluate the inference made by the LLM.}
\begin{itemize}[nosep, leftmargin=1.4em]
  \item \textbf{(a) Correct}; the inference the LLM drew is accurate.
  \item \textbf{(b) Incorrect}; the inference is not valid or is misleading.
  \item \textbf{(c) N/A}; the inference cannot be evaluated.
\end{itemize}
\end{tcolorbox}

\begin{tcolorbox}[
  colback=lightpurple,
  colframe=lightpurple,
  boxrule=0pt,
  top=2pt, bottom=4pt, left=6pt, right=6pt,
  boxsep=0pt, enlarge top by=-1pt, enlarge bottom by=0pt,
  fontupper=\small
]
\textbf{Q3: Evaluate the severity assessment made.}
Assume the LLM's inferences are correct when answering this question.
\begin{itemize}[nosep, leftmargin=1.4em]
  \item \textbf{(a) Underestimation}; the severity is assessed lower than
  warranted.
  \item \textbf{(b) Overestimation}; the severity is assessed higher than
  warranted.
  \item \textbf{(c) Within expected range}; the assessment is appropriate
  given the evidence.
\end{itemize}
\end{tcolorbox}

\end{tcolorbox}

\vspace{0.5em}
\captionof{figure}{Annotation questions and response options as presented to
annotators in the web-based annotation tool.}
\label{fig:annotation-questions}
\end{center}

\paragraph{Faithfulness Analysis of LLM Rationales: Claim Segmentation Procedure}

Prior to annotation, LLM-generated rationales were segmented by GPT-5 into individual claims to facilitate claim-level assessment of factual grounding and inference correctness. The segmentation prompt instructed the model to decompose each rationale into a sequence of discrete, self-contained observations or inferences that could be independently evaluated for hallucinations while preserving the original wording and ordering of the rationale. Final PTSD scores and score assignments were excluded from segmentation. The prompt also included manually crafted examples illustrating the desired level of granularity. The full segmentation prompt is provided below.

\begin{center}
\begin{tcolorbox}[
  enhanced,
  breakable,
  sharp corners=south,
  colframe=black,
  boxrule=0.6pt,
  colback=white,
  fonttitle=\bfseries,
  left=0pt, right=0pt, top=0pt, bottom=1pt,
  boxsep=0pt, arc=0pt
]

\begin{tcolorbox}[
  colback=lightyellow,
  colframe=lightyellow,
  boxrule=0pt,
  top=3pt, bottom=2pt, left=6pt, right=6pt,
  boxsep=0pt, enlarge top by=0mm, enlarge bottom by=-1pt,
  fontupper=\small
]
\textbf{Task:} You are given a chain-of-thought rationale generated by an AI model
that estimated a patient's PTSD severity from an interview transcript. Your task is
to segment the rationale into individual claims that can later be independently
annotated for hallucinations. The goal is to break the rationale into atomic claims
so that human annotators can determine whether each claim is grounded in the
transcript or hallucinated.
\end{tcolorbox}

\begin{tcolorbox}[
  colback=lightblue,
  colframe=lightblue,
  boxrule=0pt,
  top=2pt, bottom=2pt, left=6pt, right=6pt,
  boxsep=0pt, enlarge top by=-1pt, enlarge bottom by=-1pt,
  fontupper=\small
]
\textbf{Each claim will later be evaluated with two annotation questions:}
\begin{itemize}[nosep, leftmargin=1.4em]
  \item \textbf{Person knowledge hallucination:} Did the model make unsupported
  claims about the patient's experiences, behaviors, emotions, or symptoms?
  \item \textbf{Task knowledge hallucination:} Did the model use unsupported or
  incorrect reasoning about how symptoms or behaviors relate to PTSD severity?
\end{itemize}
\end{tcolorbox}

\begin{tcolorbox}[
  colback=lightgreen,
  colframe=lightgreen,
  boxrule=0pt,
  top=2pt, bottom=2pt, left=6pt, right=6pt,
  boxsep=0pt, enlarge top by=-1pt, enlarge bottom by=-1pt,
  fontupper=\small
]
\textbf{A claim is a single, self-contained assertion that:}
\begin{itemize}[nosep, leftmargin=1.4em]
  \item Describes a specific symptom, behavior, or psychological pattern observed
  in the patient.
  \item Expresses exactly one observation or inference. Do not bundle multiple
  unrelated rationales into a single claim.
  \item Only bundle claims together when they are parts of the same chain of
  thought of task knowledge.
\end{itemize}
\end{tcolorbox}

\begin{tcolorbox}[
  colback=lightpurple,
  colframe=lightpurple,
  boxrule=0pt,
  top=2pt, bottom=2pt, left=6pt, right=6pt,
  boxsep=0pt, enlarge top by=-1pt, enlarge bottom by=-1pt,
  fontupper=\small
]
\textbf{Guidelines:}
\begin{itemize}[nosep, leftmargin=1.4em]
  \item Keep the original wording exactly as written by the LLM.
  \item Do not include the overall diagnosis or PCL-5 score as a claim. Focus only
  on the chain of thought leading up to it.
  \item Do not include a factual rewrite of the assessment score as a claim (e.g.,
  ``In the context of the PCL score, which ranges from 17 to 85, I would estimate
  the individual's score as follows:'').
  \item Preserve ordering from the original rationale.
  \item Try to keep the number of claims to a maximum of 8.
\end{itemize}
\end{tcolorbox}

\begin{tcolorbox}[
  colback=lightgray,
  colframe=lightgray,
  boxrule=0pt,
  top=2pt, bottom=2pt, left=6pt, right=6pt,
  boxsep=0pt, enlarge top by=-1pt, enlarge bottom by=-1pt,
  fontupper=\small\ttfamily
]
\textbf{\rmfamily Output format.} Return ONLY a valid JSON array with the
following format. Do not include explanations, markdown, or additional text.
\begin{verbatim}
[
  {"id": 1, "text": "..."},
  {"id": 2, "text": "..."}
]
\end{verbatim}
\end{tcolorbox}

\begin{tcolorbox}[
  colback=white,
  colframe=lightgray,
  boxrule=0.4pt,
  top=2pt, bottom=4pt, left=6pt, right=6pt,
  boxsep=0pt, enlarge top by=-1pt, enlarge bottom by=0pt,
  fontupper=\footnotesize
]
\textbf{Example (human segmentation).}

\textit{Patient transcript:} [\,\ldots\,transcript\,\ldots\,]

\textit{LLM rationale:} ``Based on the provided transcript, I will analyze the
content, emotional tone, and references to trauma-related symptoms\ldots
The individual does not exhibit explicit PTSD symptoms, but mentions memory and
concentration issues that could be related to PTSD. The individual's emotional
tone is generally neutral\ldots''

\textit{Expected output:}
\begin{verbatim}
[
  {"id": 1, "text": "The individual's responses suggest
   significant life changes and stressors, including the
   loss of a family member, health issues, and COVID-19."},
  {"id": 2, "text": "However, the primary focus is on
   experiences related to the World Trade Center disaster."},
  {"id": 3, "text": "The individual does not exhibit explicit
   PTSD symptoms, but mentions memory and concentration
   issues that could be related to PTSD."},
  {"id": 4, "text": "The emotional tone is generally neutral,
   with some positive notes about hobbies and interests."}
]
\end{verbatim}
\end{tcolorbox}

\end{tcolorbox}

\vspace{0.5em}
\captionof{figure}{Prompt used to segment LLM chain-of-thought rationales into
individual claims for hallucination annotation.}
\label{fig:segmentation-prompt}
\end{center}

\end{document}